\RequirePackage{fix-cm}
\documentclass[twocolumn]{svjour3}

\usepackage{natbib}
\usepackage{graphicx}%
\usepackage{multirow}%
\usepackage{amsmath,amssymb,amsfonts}%
\usepackage{mathrsfs}%
\usepackage[title]{appendix}%
\usepackage{xcolor}%
\usepackage{textcomp}%
\usepackage{manyfoot}%
\usepackage{booktabs}%
\usepackage{algorithm}%
\usepackage{algorithmicx}%
\usepackage{algpseudocode}%
\usepackage{listings}%
\usepackage{colortbl}
\usepackage{anyfontsize}
\usepackage{pifont}
\usepackage[most]{tcolorbox}

\usepackage[USenglish]{babel}
\usepackage{hyperref} 
\hypersetup{breaklinks=true,colorlinks,citecolor=black}

\newcommand{\be}{\begin{eqnarray}}
\newcommand{\ee}{\end{eqnarray}}
\newcommand{\bee}{\begin{eqnarray*}}
\newcommand{\eee}{\end{eqnarray*}}

\newcommand{\matrixb}{\left[ \begin{array}}
\newcommand{\matrixe}{\end{array} \right]}

\newcommand{\app}{\raise.17ex\hbox{$\scriptstyle\sim$}}
\input{def_custom.tex}

\begin{document}
\sloppy

\title{Hearing and Seeing Through CLIP: A Framework for Self-Supervised Sound Source Localization
%

}


\author{ Sooyoung Park*  \and
         Arda Senocak* \and
         Joon Son Chung
}


\institute{
              *These authors contributed equally to this work.\\
              Sooyoung Park \at
              School of Electrical Engineering, KAIST, South Korea \\
              ETRI, South Korea \\
              \email{sooyoung@etri.re.kr}
           \and
           Arda Senocak \at
              School of Electrical Engineering, KAIST, South Korea \\
              \email{arda.senocak@gmail.com}
           \and
           Joon Son Chung \at
               School of Electrical Engineering, KAIST, South Korea \\
              \email{joonsc@kaist.ac.kr}
}

\date{Received: date / Accepted: date}

\maketitle

\abstract{
Large-scale vision-language models demonstrate strong multimodal alignment and generalization across diverse tasks. Among them, CLIP stands out as one of the most successful approaches. In this work, we extend the application of CLIP to sound source localization, proposing a self-supervised method operates without explicit text input. We introduce a framework that maps audios into tokens compatible with CLIP’s text encoder, producing audio-driven embeddings. These embeddings are used to generate sounding region masks, from which visual features are extracted and aligned with the audio embeddings through a contrastive audio-visual correspondence objective. Our findings show that alignment knowledge of pre-trained multimodal foundation model enables our method to generate more complete and compact localization for sounding objects. We further propose an LLM-guided extension that distills object-aware audio-visual scene understanding into the model during training to enhance alignment. Extensive experiments across five diverse tasks demonstrate that our method, in all variants, outperforms state-of-the-art approaches and achieves strong generalization in zero-shot settings.
\keywords{Audio-visual Learning \and Sound Source Localization \and Discriminative Learning}
}

\maketitle

\section{Introduction.} 
\label{sec:intro}

\begin{figure}[t]
\centering
        \includegraphics[width=1.0\linewidth]{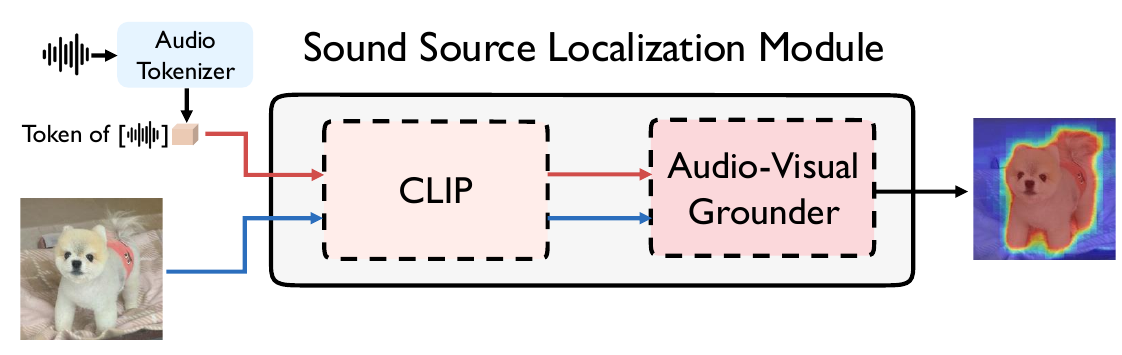}
\caption{\textbf{The proposed CLIP based sound source localization method.} 
}
\label{fig:teaser}
\end{figure}
Localizing sound sources is a fundamental aspect of human and animal perception, allowing us to understand and navigate complex environments by integrating auditory and visual cues. Inspired by this capability, audio-visual sound source localization has gained significant attention in recent years~\citep{senocak2018learning, arandjelovic2018objects, senocak2019learning, qian2020multiple, hu2020discriminative, chen2021localizing, lin2021unsupervised, li2021space, senocakLessMore, song2022sspl, senocakHardPos, ssslTransformation, ezvsl, marginnce, slavc, sun2023learning, senocak2023alignment}. A main challenge in this field is enabling machines to identify which visual entities are producing sounds without explicit supervision. To tackle this, many approaches leverage the natural semantic alignment between audio and visual modalities. Among them, self-supervised contrastive learning has emerged as a dominant paradigm, aligning representations of paired audio and visual inputs to learn robust, generalizable cross-modal embeddings without the need for labeled data.

While many sound source localization methods are built on the principle of learning audio-visual semantic correspondence, some incorporate additional priors to guide localization. These include visual objectness cues~\citep{ezvsl, slavc}, object proposal networks~\citep{xuan2022proposal}, and auxiliary modalities such as optical flow~\citep{htf}. Although such heuristics may help localization performance, they often fail to improve true semantic alignment~\citep{senocak2023alignment, senocak2024aligning} and can introduce biases, leading models to exploit shortcuts that undermine meaningful cross-modal reasoning~\citep{oya2020we, slavc, arandjelovic2018objects, senocak2023alignment, senocak2024aligning}. In contrast, our approach enhances audio-visual alignment by leveraging strong cross-modal priors learned from large-scale multimodal data. We employ the CLIP model~\citep{radford2021learning}, known for its robust alignment of visual content with natural language and its rich, flexible embedding space that encodes broad semantic information across modalities with strong descriptive power. We build our self-supervised sound source localization method on top of this pretrained alignment knowledge.

Most frameworks that leverage the CLIP model incorporate text prompts. However, in this work, we explore an approach that does not rely on explicit contextual text input. This decision is motivated by several key observations: (1) sound source localization datasets, such as SoundNet-Flickr, do not provide paired textual annotations; (2) the task itself is inherently unlabeled and audio-driven; and (3) a principled approach to sound source localization should rely on learning semantic alignment between audio and visual modalities through self-supervision, without requiring class labels. Methods that depend on class-label-derived text inputs are not universally applicable, as such labels are often unavailable in real-world, in-the-wild scenarios. Therefore, we adopt a core framework that employs the pre-trained CLIP model in a configuration that does not use class-label textual input, but instead relies solely on audio-driven inputs (as illustrated in~\Fref{fig:teaser}), using audio-visual correspondence as the supervisory signal.

 We propose the following pipeline as our core framework: First, we translate audio signals into tokens compatible with CLIP’s text encoder, producing contextual audio-driven embeddings. These embeddings are then aligned with visual features using contrastive learning. Specifically, we use the audio-driven embeddings to highlight sounding regions in the visual scene and extract visual features from those regions at both the image and feature levels. These visual features are subsequently aligned with the audio-driven embeddings through an audio-visual correspondence objective. The entire model is trained end-to-end within this self-supervised contrastive learning framework.

We extensively evaluate our method across diverse set of sound source localization tasks -- including single-source localization, audio-visual robustness, segmentation, interactive localization, and multi-source localization -- on various datasets. Our experiments consistently show that the proposed approach outperforms existing methods by a significant margin across all benchmarks. These results highlight the potential of our method as a generalizable model in zero-shot settings.

As an extension of our core framework, we propose an alternative learning objective that enhances audio-visual correspondence by distilling object-aware scene understanding from an Large Language Model (LLM) via CLIP’s text encoder, used only during training. Built on CLIP, our framework remains versatile and can incorporate text embeddings when auxiliary text inputs are available. This variant is designed to operate automatically on in-the-wild samples without requiring ground-truth class labels or manually annotated captions, making it broadly applicable to real-world scenarios where labeled metadata is unavailable. This alternative approach further improves our state-of-the-art results across various tasks.

We summarize our contributions as follows:
\begin{itemize}
    \item We propose a self-supervised sound source localization framework that leverages CLIP without requiring direct textual prompts or class labels.
    \item We introduce an AudioTokenizer module that maps audio into tokens compatible with CLIP’s text encoder, enabling effective audio-visual alignment.
    \item Our method achieves strong zero-shot generalization across five diverse tasks and consistently outperforms existing approaches.
    \item We extend our framework with an LLM-based training objective that distills object-aware audio-visual scene understanding to further boost performance.
\end{itemize}
\section{Related work}\label{sec:RW}
\newpara{Sound source localization.} Audio-visual sound source localization aims to detect sound-emitting regions, objects, or events within a visual scene by understanding the semantic correspondences between audio and visual modalities. Initial studies~\citep{senocak2018learning,arandjelovic2018objects,senocak2019learning} investigated this problem by employing techniques such as cross-modal attention and contrastive learning to achieve effective alignment between the two modalities. However, in real-world scenarios, audio-visual pairs may be imperfectly aligned due to factors such as background noise, off-screen sound sources, or silent visual objects, all of which can introduce misleading associations and hinder accurate localization. To mitigate these challenges, previous approaches have proposed techniques such as false negative-aware learning~\citep{sun2023learning}, negative-free predictive learning~\citep{song2022sspl}, and various regularization strategies in the contrastive learning formulation~\citep{marginnce, slavc}. Another line of research in self-supervised contrastive learning focuses on enhancing audio-visual alignment by leveraging training data through effective strategies, such as sample mining~\citep{senocakHardPos, chen2021localizing}, geometric equivalence learning~\citep{ssslTransformation}, and multi-positive contrastive learning~\citep{senocak2023alignment,senocak2024aligning}. Furthermore, several sound localization methods explore using additional prior knowledge or post-processing techniques. For example,~\citep{senocakLessMore, qian2020multiple} incorporate label information to supervise the training of backbone audio and visual networks or to enhance audio-visual alignment. Xuan~\etal\citep{xuan2022proposal} use object priors in the form of object proposals, while Mo~\etal~\citep{ezvsl} apply a post-processing that refines localization outputs using pre-trained visual feature activation maps. In contrast, our approach -- while adopting a fully self-supervised framework like existing methods -- leverages CLIP’s multimodal alignment knowledge as a prior to achieve robust audio-visual alignment.

Another important research direction in visual sound source localization is multi-source sound localization~\citep{hu2020discriminative,qian2020multiple,hu2022mix,mo2023audio,mahmud2024t,kim2024learning}, where multiple sound-emitting objects are present within a scene. These works primarily focus on localizing generic sound sources. More recent studies~\citep{hamilton2024separating,ryu2025seeing} extend this direction by aiming to localize both generic sounds and spoken utterances from audio mixtures. In contrast, our method is designed for single sound source localization and is trained specifically for that objective. Nonetheless, due to its strong audio-visual alignment capabilities, it can be directly applied to multi-source generic sound localization tasks too.

\newpara{CLIP in Audio-Visual Learning.} Recent contrastive language-image pretraining (CLIP) models, which are pretrained on large-scale paired data~\citep{radford2021learning,jia2021scaling}, demonstrate robust generalization ability and have been successfully used in numerous downstream tasks across various research topics. In this section, we review related works that incorporate CLIP~\citep{radford2021learning} for audio-visual learning. WAV2CLIP~\citep{wu2022wav2clip} and AudioCLIP~\citep{guzhov2022audioclip} expand the pre-trained CLIP model by aligning audio features with text and visual features in a shared embedding space, \ie representation learning. They achieve this either using paired data or by utilizing the visual modality as a bridge. Beyond representation learning, CLIP models are also employed in audio-visual event localization~\citep{mahmud2023ave} and video parsing~\citep{fan2023revisit}, as well as audio-visual source separation~\citep{tan2023language,dong2022clipsep}. While~\citep{tan2023language} employs text input for separation, CLIPSep~\citep{dong2022clipsep} is trained based on the audio-visual relationship without text. Similarly, our proposed method is also trained solely with an audio-visual alignment objective. Another line of work~\citep{yariv2023audiotoken,bhatisegmental} adapt pre-trained CLIP models and text encoders for audio. They achieve this by mimicking contextual text tokens using audio signals, enabling the CLIP text encoder to embed audio signals. Our work also employs a similar approach to leverage the CLIP model with audio-driven input for the sound localization task.

More recently, T-VSL~\citep{mahmud2024t} proposed a text-guided multi-source sound localization framework based on the tri-modal joint embedding model, AudioCLIP. While their work also addresses sound source localization, our approach differs fundamentally in its design. Instead of relying on AudioCLIP, which operates within a tri-modal embedding space, we build upon the original CLIP model which does not include an audio modality. To use audio, we introduce a novel AudioTokenizer module that effectively mimics CLIP’s textual embedding space. This design enables our model to directly align audio embeddings with visual features, without requiring explicit text embeddings during either training or inference.

This work extends~\citep{park2024can} by introducing an enhanced training framework with an LLM-based objective that distills object-aware audio-visual scene understanding, conducting extensive experiments across five sound localization tasks to demonstrate the method’s generalizability in zero-shot settings, and providing analysis showing AudioTokenizer enables learning semantically descriptive audio embeddings.
\begin{figure*}[tp]
    \centering
    \includegraphics[width=0.95\linewidth]{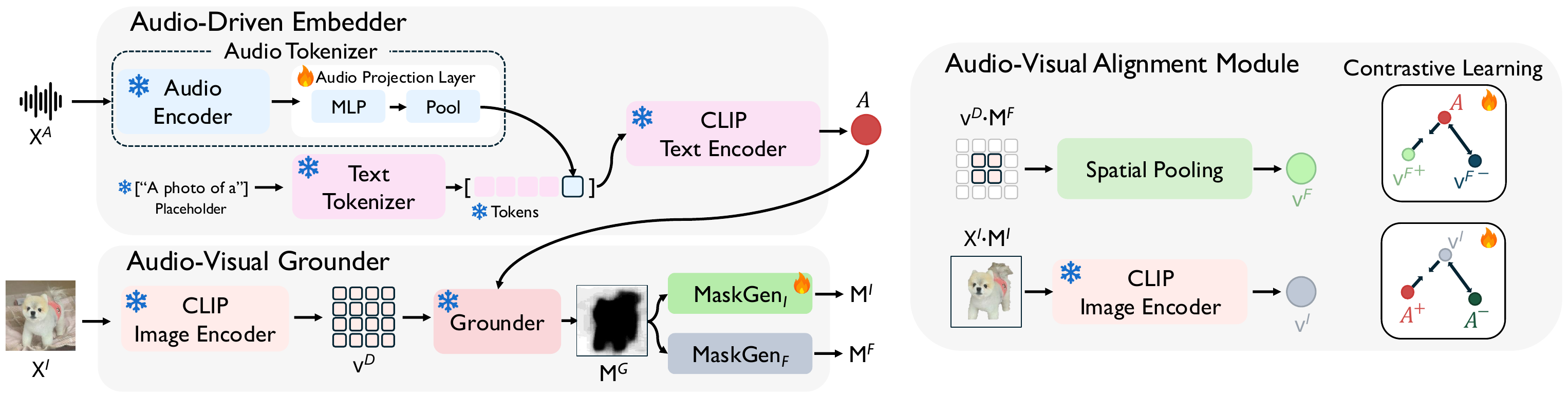}
    \caption{{\bf Our sound source localization framework with Audio-Visual Alignment.} The proposed method takes audio-visual pairs, translating audio signals into CLIP-compatible tokens via the Audio Tokenizer module to generate audio-driven embedding, $\mathbf{A}$. This embedding highlights sounding regions within the Audio-Visual Grounder module. With the sounding area masks, the Audio-Visual Alignment module extracts audio-grounded visual features at both image-level ($\boldsymbol{v}^I$) and feature-level ($\boldsymbol{v}^F$). These visual features and audio feature are aligned via contrastive learning.}
    \label{fig:pipeline}
\end{figure*}
\section{Method}\label{sec:MTD}
\subsection{Audio-Driven Embedder} \label{ssec:tokenizer}
Our goal is to use the CLIP text encoder to embed audio without requiring any text input. To this end, we introduce the \textbf{AudioTokenizer} module, which transforms audio into text-like tokens that can be directly processed by CLIP’s pre-trained text encoder. Conceptually, the module translates audios into discrete embeddings that mimic natural language tokens. These `audio tokens' serve as a bridge between the audio modality and CLIP's language space, enabling us to leverage CLIP’s powerful multimodal alignment capabilities without relying on explicit textual annotations.

The \textbf{AudioTokenizer} consists of two main components: an \textbf{audio encoder} and a \textbf{projection network}. The audio encoder, $E_{A}$, is a transformer-based model pre-trained in a self-supervised manner, following~\citep{chen2022beats}. It takes a spectrogram and produces audio embeddings that capture the semantic content of the sound. These embeddings are then passed through a lightweight projection network, composed of two MLP layers and an attentive pooling layer (as in~\citep{yariv2023audiotoken}), which transforms them into a form compatible with CLIP’s token embeddings. While the audio encoder is pre-trained and kept frozen during training, the remaining layers are trained end-to-end in our sound source localization framework, with the objective of audio-visual alignment (see Eq.~\ref{eq:final} or Eq~\ref{eq:final_caption}).

The resulting `audio token' serves as a semantic representation of the audio and is treated as a substitute for a word token. It is appended to a fixed textual prompt prefix, such as ``A photo of a'', forming a pseudo-text sequence as in~\Fref{fig:pipeline}. This hybrid token sequence is then passed through CLIP’s frozen text encoder, $E_{\text{CLIP}t}$, to produce a final audio-driven embedding, $\mathbf{A}$. Importantly, because CLIP’s text and vision encoders are jointly aligned during pretraining, the audio-driven embedding $\mathbf{A}$ inherits visual alignment properties by virtue of being processed through $E_{\text{CLIP}_t}$. This design enables seamless pairing or conditioning of $\mathbf{A}$ with any CLIP-based image encoder.
\vspace{-4mm}
\subsection{Audio-Visual Grounder} \label{ssec:grounder}
After extracting the audio-driven embeddings, we also extract visual features of each audio-visual pair. These features are then passed to our audio-visual grounder, which performs grounding by identifying regions in the image associated with the sound and generating corresponding masks. These masks are subsequently used to extract visual embeddings at both the image-level and the feature-level, which are then utilized in the audio-visual alignment objective. Our Audio-Visual grounding module is designed with three components: 1) an image encoder, 2) a grounder, and 3) mask generators.

We use a pre-trained CLIP image encoder as our image encoder, denoted as $E_{\text{CLIP}_v}$. It is responsible for encoding the provided input images into spatial features. For our grounder, $G$, we employ off-the-shelf CLIP-based segmentation network known as CLIPSeg~\citep{luddecke2022image}. It is important to note that CLIPSeg requires CLIP-based visual features and text conditioning to perform segmentation. We leverage the outputs from our image encoder as visual features for grounder. However, since our approach does not use any text input directly, we utilize our audio-driven embedding, $\mathbf{A}$, for conditioning. The result of the grounder $G$, $\mathbf{M}^G$, is potential sounding regions. Both the image encoder and the grounder remain fixed during training.

As a common approach, sound source localization methods use audio-attended visual embeddings alongside audio embeddings within the audio-visual alignment objective. Therefore, it is essential for our method to generate differentiable binary masks that accurately represent the sounding regions. We introduce two masking methods: Image-level Mask Generator ($MaskGen_I$) and Feature-level Mask Generator ($MasGen_F$), both of which serve to extract audio-grounded visual embeddings at the image and feature levels, respectively. Similar to~\citep{cha2023learning}, $MaskGen_I$ utilizes a learnable scalar projection $(w\cdot \mathbf{M}^G+b)$ on the output of the grounder, $\mathbf{M}^G$, and then applies the Gumbel-Max technique~\citep{jang2016categorical} to generate a differentiable binary mask, referred to as $\mathbf{M}^I$. This mask is used to identify sounding areas in the image. $MaskGen_F$ is designed with min-max normalization and soft-thresholding functions applied to $\mathbf{M}^G$ to obtain $\mathbf{M}^F$, which allows the extraction of audio-visually correlated areas at the feature level. The utilization of these mask generators is explained in the following section.
\vspace{-6mm}
\subsection{Alignment Module} \label{ssec:alignment}
After obtaining sounding region masks for the given audio-visual pairs from the audio-visual grounder, our method extracts contextual visual embeddings from the masked areas at both image and feature levels, $\boldsymbol{v}^I$ and $\boldsymbol{v}^F$ respectively. These visual embeddings are then aligned with the audio-driven embedding, $\mathbf{A}$, as part of the audio-visual alignment objective. For this purpose, we define two contrastive learning losses: image-level and feature-level audio-grounded contrastive losses, $ACL_I$ and $ACL_F$. In a nutshell, our model learns to maximize the alignment between the visual features of sounding regions and the corresponding audio features.

\newpara{Image-Level Audio-Grounded Contrastive Loss.} Different from typical global image and audio correspondence, our focus is on alignment between sounding region and audio. One approach to achieve this is by highlighting the sounding regions (foreground pixels) in the image and masking out the background areas, as depicted in~\Fref{fig:pipeline}. To begin, the mask $\textbf{M}_i^I$ obtained from $MaskGen_I$ for the $i$th audio-visual pair is used to mask out the irrelevant areas in the image.
This masked image is then transformed into a visual embedding by using CLIP image encoder $E_{\text{CLIP}_v}$, ${\boldsymbol{v}}^I_i= E_{\text{CLIP}_v}\left( \textbf{M}_i^I\cdot \textbf{X}^I_i \right)$. The audio-visual similarity between the audio-driven embedding $\mathbf{A}_i$ and the audio-grounded visual embedding $\boldsymbol{v}_i^I$ is computed using cosine similarity and defined as $S^I_{i,i} = ({{\boldsymbol v}_i^I}\cdot \mathbf A_i)$. We employ symmetric InfoNCE for the contrastive loss. We note that image-level masks are computed only for positive pairs. Thus, the objective of this loss is to maximize the similarity between the positive sounding region and the corresponding audio pair, while also ensuring dissimilarity between negative audios and the actual sounding region. The $ACL_I$ loss is defined as follows:
\begin{align}
    \mathcal L_{ACL_I} = & \; \textit{InfoNCE}(\textbf{S}^I) \nonumber \\
    = &-\frac1{2B} \sum^B_i \log \frac{\exp(S^I_{i,i}/\tau)}{\sum^B_j \exp(S^I_{i,j}/\tau)} \nonumber \\
    & - \frac1{2B} \sum^B_i \log \frac{\exp(S^I_{i,i}/\tau)}{\sum^B_j \exp(S^I_{j,i}/\tau)}
    \label{eq:acli}
\end{align}

\noindent where $\tau$ is the temperature and $\textbf{S}^I$ is image-level audio-visual similarity matrix within batch. With the help of this loss, the sounding region and the generated mask $\textbf{M}^I$ gradually cover the target sounding area. However, we observe that $ACL_I$ alone can not enable the model to completely suppress the background regions.

\begin{figure*}[tp]
    \centering
    \includegraphics[width=0.95\linewidth]{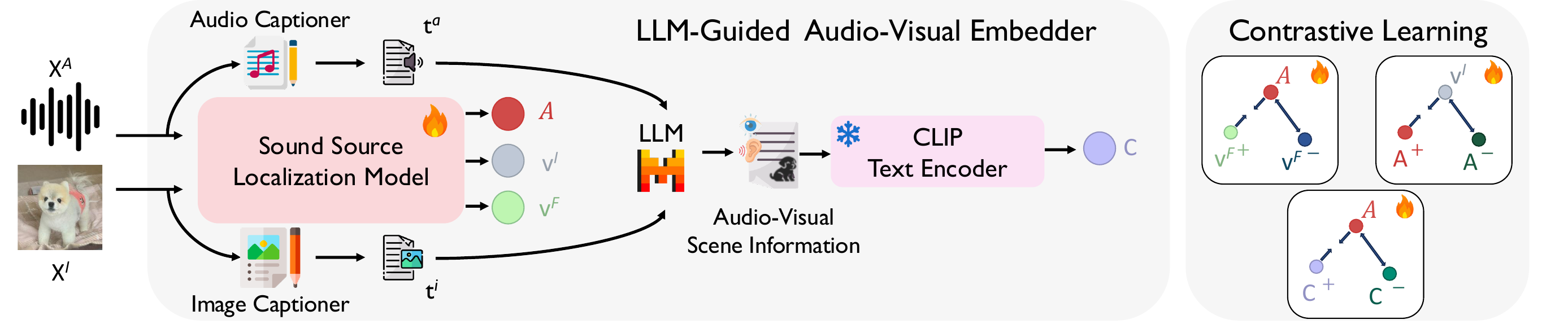}
    \caption{{\bf Our extended framework with  LLM-Guided Object-Aware Alignment.} The proposed alternative method extends our audio-visual alignment framework by incorporating LLM guidance. Given audio-visual pairs, captions from each modality, $\mathbf{t}^a$ and $\mathbf{t}^i$, are generated and processed by an LLM to extract object-aware scene information. This is then encoded using CLIP’s text encoder, $\mathbf{c}$, and aligned with the audio-driven embedding $\mathbf{A}$, via contrastive learning, serving as an auxiliary objective. Since CLIP’s text and visual features are aligned, this guidance implicitly reinforces audio-visual correspondence.}
    \label{fig:llm_pipeline}
\end{figure*}

\newpara{Feature-Level Audio-Grounded Contrastive Loss.}
Suppressing masks derived from negative pairs is essential for enhancing robustness against background regions. However, due to memory constraints, generating high-resolution image-level masks for all negative pair combinations within a batch is infeasible. As an alternative, we introduce the feature-level audio-grounded contrastive loss, $ACL_F$, allowing the use of masks in lower-resolution (on features), effectively bypassing the memory constraints. This strategic approach involves emphasizing regions within the spatial visual features, as shown in Figure \ref{fig:pipeline}. To elaborate, the mask $\mathbf{M}^F_{i,j} \in \mathbb{R}^{h \times w}$ obtained from the $MaskGen_F$ for given image $\textbf{X}^I_i$ and audio $\textbf{X}^A_j$, is applied during spatial pooling of the spatial visual features $\boldsymbol{v}^D_i \in \mathbb{R}^{c \times h \times w}$ to focus on regions within the features that exhibit high correlation with the paired audio. Feature-level audio-grounded visual embedding $\boldsymbol{v}^F_{i,j} \in \mathbb{R}^{c}$ is as follows:
\begin{equation}
    \boldsymbol{v}^F_{i,j} = \frac{\sum_{h,w}\mathbf{M}^F_{i,j,h,w} \cdot \boldsymbol{v}^D_{i,h,w}}{\sum_{h,w}\mathbf{M}^F_{i,j,h,w}}.
        \label{eq:pooling}
\end{equation}
In contrast to $ACL_I$, which focuses on the sounding region, $ACL_F$ focuses on the highly correlated area, regardless of positive or negative audio-visual pairs. The audio-visual similarity between the audio-driven embedding $\mathbf{A}$ and the feature-level audio-grounded visual embedding $\boldsymbol{v}^F$ for both positive and negative pairs is computed using cosine similarity defined as $S^F_{i,j}=( \boldsymbol{v}^F_{i,j} \cdot \textbf{A}_j )$. The $ACL_F$ loss is defined as follows:
\begin{align}
    \mathcal L_{ACL_F} = \textit{InfoNCE}(\textbf{S}^F),
    \label{eq:aclf}
\end{align}
\noindent where $\textbf{S}^F$ is feature-level audio-visual similarity matrix within batch. One may question the necessity of introducing a separate mask generator, $\textbf{M}^F$, when $\textbf{M}^I$ already exists. While it is technically possible to reuse the image-level mask $\textbf{M}^I$ in Eq.~\ref{eq:pooling}, doing so can lead to unintended training dynamics. Specifically, $\textbf{M}^I$ may produce masks that are nearly zero-valued when processing negative audio-visual pairs. This behavior can cause the numerator in Eq.~\ref{eq:pooling} effectively being zero, resulting in $\boldsymbol{v}^F_{i,j}$ becoming arbitrary or uninformative. Consequently, this can degrade the contrastive training process by producing random or unstable similarity scores in the InfoNCE loss for negative pairs. To mitigate this issue and maintain robust training, we decouple the mask generators and define separate modules for image-level and feature-level masking.

\subsection{Area Regularization} \label{ssec:area}
We observe that even when training with the $ACL_I$ and $ACL_F$ losses, the model may adopt a shortcut by generating masks that include both relevant and irrelevant regions -- sometimes covering the entire image. This behavior arises because the CLIP image encoder in the Alignment module can still produce semantically meaningful visual features, even when the masked region is overly broad. For example, the entire image of a dog, a loosely cropped foreground, or an accurately segmented dog image can all yield similar high-level CLIP features corresponding to the `dog' concept. As a result, the model receives a weak training signal and may not be incentivized to localize precisely. To address this, and following the regularization strategies proposed in~\citep{xie2022clims,cha2023learning}, we introduce an area regularization loss, defined as follows:
\begin{equation}
	\label{eq:Reg}
	\mathcal{L}_{Reg} = \sum_{i} \Vert p^+ - \overline{\mathbf{M}^I_{i,i}} \Vert_1 + \sum_{i \ne j} \Vert p^- - \overline{\mathbf{M}^I_{i,j}} \Vert_1,
\end{equation}

\noindent where $\mathbf{M}^I_{i,i}$ and $\mathbf{M}^I_{i,j}$ are the image masks from the positive and negative pairs respectively. The area of these masks are denoted as $\overline{\mathbf{M}}$. $p^+$ and $p^-$ represent the area prior hyperparameters, set to 0.4 and 0.0. The area regularizer constrains the size of the mask during learning to ensure that the intended sounding regions are contained while irrelevant areas are discarded. 
\vspace{-6mm}
\subsection{Training} \label{ssec:training}
The overall training loss term is defined as follows:
\begin{equation}
	\label{eq:final}
	\mathcal{L} = \lambda_{ACL_I} \mathcal{L}_{ACL_I} + \lambda_{ACL_F} \mathcal{L}_{ACL_F} + \lambda_{REG} \mathcal{L}_{REG},
\end{equation}
where $\lambda_{ACL_I}$, $\lambda_{ACL_F}$, and $\lambda_{REG}$ are the hyper-parameters weighting the loss terms.

\subsection{LLM-Guided Object-Aware Alignment} \label{ssec:new_method}
In this section, we introduce an alternative learning approach to train our model by leveraging the audio-visual scene understanding capabilities of Large Language Models (LLMs). As demonstrated in~\citep{sung2024avhbench}, recent LLMs have the ability to disentangle audio-visual information from metadata that describe a scene. Inspired by this, we aim to distill the object-aware audio-visual scene understanding of an LLM to enhance the audio-visual correspondence objective of our proposed method. To provide fine-grained descriptions of the scene to the LLM, we utilize off-the-shelf captioning models for each modality. For each given audio-image pair, the samples are fed into their respective captioning models, producing one caption per modality. These two captions are then provided to the LLM to obtain an object-aware understanding of the audio-visual scene. We design our LLM prompt as follows for a given image and audio captions:
\begin{tcolorbox}[colback=gray!10,colframe=gray!10,sharp corners]
Identify the primary object in the `image caption' most likely producing the sound like given `audio caption', excluding background sounds which is hard to infer from given `image caption'. Keep the answer concise and focused on general concepts, such as type. Limit the response to no more than 3 words.

Image caption: \{...\},
Audio caption: \{...\}

\end{tcolorbox}
We can formulate entire object-aware audio-visual scene knowledge stage of our pipeline as follows: $\boldsymbol{\mathcal{T}}_{\text{image}} = \{ t_i = G_{\text{I2T}}(\mathbf{x}_i)) \mid \forall \mathbf{x}_i \in \boldsymbol{\mathcal{D}} \}$ and $\boldsymbol{\mathcal{T}}_{\text{audio}} = \{ t_a = G_{\text{A2T}}(\mathbf{x}_i)) \mid \forall \mathbf{x}_i \in \boldsymbol{\mathcal{D}} \}$, where $\boldsymbol{\mathcal{D}} = \{ \mathbf{x}_1, \mathbf{x}_2, \dots, \mathbf{x}_N \}$ is the dataset with $N$ samples (audio-visual pairs), and $G_{\text{I2T}}$ and $G_{\text{A2T}}$ represent the captioners for the image and audio modalities, respectively. Then, object-aware audio-visual scene information is obtained as $\boldsymbol{\mathcal{C}}_{\text{object}} = \{ c_i = LLM(p,t_i,t_a) \mid \forall t_i \in \boldsymbol{\mathcal{T}}_{\text{image}}, \forall t_a \in \boldsymbol{\mathcal{T}}_{\text{audio}} \}$.

As our proposed framework is built upon CLIP, it remains versatile, allowing the integration of CLIP’s text encoder and its embeddings whenever text input is available. After the LLM generates its object-aware audio-visual scene understanding, $c_i$, we encode this information using CLIP’s text encoder, $E_{\text{CLIP}_t}(c_i)$, to obtain an additional supervisory signal (\Fref{fig:llm_pipeline}). This representation is used alongside our original audio-visual alignment objective during training. Since CLIP’s text and visual embeddings are already well aligned within the same embedding space, aligning our audio-driven embeddings to the CLIP text features inherently brings them closer to the visual features as well. By introducing this additional alignment path, we aim to strengthen the robustness of the audio-visual correspondence learning, ultimately helping the audio features better align with the visual modality through the shared CLIP space.

To explicitly incorporate this additional alignment into our training objective, we define a caption-audio contrastive loss, $ACL_C$. For each sample, the audio-caption similarity $S^C_{i,j} = ({{\boldsymbol C}_i} \cdot \mathbf{A}j)$ is calculated via cosine similarity between the caption embeddings, ${{\boldsymbol C}_i} = E_{\text{CLIP}_t}(c_i)$, and audio embeddings. Following the same InfoNCE formulation used in previous alignment objectives, the caption-level contrastive loss is defined as:

\begin{equation} 
    \mathcal{L}_{ACL_C} = \textit{InfoNCE}(\textbf{S}^C), 
\end{equation} 
where $\textbf{S}^C$ denotes the caption-audio similarity matrix within a batch. The total loss is formulated by adding the audio-caption similarity loss, $\mathcal{L}_{ACL_C}$, to the overall training loss term defined in Eq.~\ref{eq:final}, as follows:

\begin{align}
\label{eq:final_caption}
\mathcal{L} =\ & \lambda_{ACL_I} \mathcal{L}_{ACL_I} + \lambda_{ACL_F} \mathcal{L}_{ACL_F} + \lambda_{REG} \mathcal{L}_{REG} \nonumber \\
& + \lambda_{ACL_C} \mathcal{L}_{ACL_C}.
\end{align}

It is important to emphasize that sound source localization is an unlabeled, audio-input-driven task, where ground-truth class labels are often unavailable across datasets. Consequently, methods that rely on text inputs derived from class labels are not universally applicable and can only be used in datasets where such annotations exist. In contrast, our proposed variant is designed to operate on any in-the-wild sample, as it does not require access to ground-truth class categories or manually annotated captions. Instead, object-aware audio-visual scene understanding is distilled through auxiliary encoders, rather than relying directly on ground-truth text labels. This design choice ensures that our method remains flexible and broadly applicable across diverse real-world scenarios where labeled metadata is not available.

\subsection{Inference} \label{ssec:inference}
For the provided image and audio pairs, an audio-driven embedding is acquired and fed into the grounder $G$ along with the visual features obtained from the image encoder. The resulting output of the grounder, $\mathbf{M}^G$, is subsequently used in $MaskGen_I$. Unlike training, during inference, it is adjusted using $\sigma\left(\mathbf{M}^G+ {b}/{w}\right)$, where $w$, $b$ are scalar projection parameters learned during training in the image masker $MaskGen_I$ and $\sigma$ is sigmoid function. The final output mask is then used to obtain the localization result. Note that, regardless of the variant, our model uses only the given image and audio pairs during inference, without requiring any additional input.
\section{Experiments}
\subsection{Experimental Settings}\label{ssec:exp_settings}
\newpara{Datasets.} 
Our model is trained on the VGGSound dataset~\citep{VGGSound}, which contains~\app 200K videos. After training, we evaluate its performance on the following datasets for single sound source localization: VGG-SS~\citep{chen2021localizing}, SoundNet-Flickr Test~\citep{senocak2018learning,senocak2019learning}, IS3~\citep{senocak2024aligning}, VPO-SS~\citep{chen2024unraveling}, VPO-MS~\citep{chen2024unraveling}, AVSBench-S4~\citep{zhou2022avs}, and DenseAV ADE20K~\citep{hamilton2024separating}. Similarly, IS3, VPO-SS, VPO-MS, AVSBench-S4, AVSBench-MS3~\citep{zhou2022avs}, and DenseAV ADE20K are used for audio-visual segmentation. In addition, IS3 and VPO-MS are employed for evaluating interactive sound source localization. These evaluation perspectives and datasets are adopted from~\citep{senocak2024aligning}. Extended VGG-SS and SoundNet-Flickr datasets proposed by~\citep{slavc} are used for audio-visual robustness. Finally, VGGSound-Instruments~\citep{hu2022mix} and VGGSound-Duet~\citep{mo2023audio} datasets are used for multisource sound localization.

\newpara{Evaluation Metrics.} 
Following the evaluation protocols in~\citep{senocak2024aligning,slavc,zhou2022avs,hu2022mix}, we utilize a set of metrics for each task: for single sound source localization, we use cIoU, cIoU Adaptive, AUC, and AUC Adaptive; for audio-visual robustness, the metrics are AP, max-F1, and LocAcc; for audio-visual segmentation, we use mIoU, mIoU Adaptive, F-Score, and F-Score Adaptive; for interactive localization, the metrics are IIoU, IIoU Adaptive, IAUC, and IAUC Adaptive; and for multisource localization, we use CAP, PIAP, AUC and cIoU.

\newpara{Implementation details.} We employ frozen pre-trained ``ViT-B/16'' CLIP~\citep{radford2021learning} model as image encoder, BEATs~\citep{chen2022beats} for audio encoder and CLIPSeg~\citep{luddecke2022image} for grounder. In the LLM-guided variant, we use BLIP-2~\citep{li2023blip} and GAMA~\citep{ghosh2024gama} as the vision and audio captioners, respectively, and Mistral-7B-Instruct-v0.2~\citep{Jiang2023Mistral7} as the language model. During training, we used 10-second audio segments sampled at 16kHz, and the center frame of the video resized to 352x352. For the overall loss, we set the parameters $\lambda_{ACL_I}$, $\lambda_{ACL_F}$, $\lambda_{ACL_C}$, and $\lambda_{Reg}$ all to 1. Additionally, we used $\tau$ as 0.07 in Equation \ref{eq:acli}. The model is optimized for 20 epochs with a batch size of 16, using the Adam optimizer with a learning rate of $10^{-4}$ and a weight decay of $10^{-4}$.

\newpara{Baselines in Quantitative Comparisons.} Besides the existing works, we also compare our proposed method with closely-related baselines that can be obtained using different components of our overall architecture. Baseline details are below:
\begin{itemize}
    \item \includegraphics[height=1em]{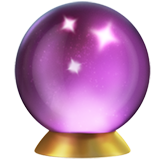}\textbf{CLIPSeg (w/ GT Text).} 
    We utilize the ground truth class labels of test samples as text conditions to obtain the segmentation results from CLIPSeg, essentially serves as an \emph{Oracle} method.
    \item \textbf{CLIPSeg (w/ WAV2CLIP Text).} WAV2CLIP aligns text, vision, and audio embeddings together in the CLIP space. For a given audio, the most relevant text (class label) can be retrieved. This retrieved text is used with CLIPSeg to highlight the sounding region in the image.
    \item \textbf{CLIPSeg (w/ CLAP Text).} CLAP~\citep{wu2023large} is designed to learn audio representations by aligning them with natural language descriptions through contrastive learning, enabling models to perform tasks such as zero-shot audio classification and audio-text retrieval. Similarly, as described above, given an audio sample, the most relevant text (class label) can be retrieved and used with CLIPSeg to obtain the grounded visual region. Note that the three baselines mentioned above can only be applied when ground-truth class labels are available, and therefore cannot be used universally.
    \item \textbf{WAV2CLIP and AudioCLIP.} These models leverage the pre-trained CLIP model to align text, vision, and audio embeddings. To enable zero-shot sound source localization with these models, we utilize a pre-trained CLIP-like object detector~\citep{li2022glip} to extract region proposals from the images and calculate the cosine similarity between the visual features of those regions and the audio features. The region with the highest similarity is employed as the localization result.
\end{itemize}

\begin{table*}
    \centering
    \resizebox{1.0\linewidth}{!}{
    \begin{tabular}{l|cccc|cccc}
    \toprule
    & \multicolumn{4}{c}{\textbf{VGG-SS}} & \multicolumn{4}{c}{\textbf{Flickr-SoundNet}} \\
    \textbf{Method} &  \textbf{cIoU $\uparrow$} &  \textbf{w/ Adap. $\uparrow$} & \textbf{AUC $\uparrow$} & \textbf{w/ Adap. $\uparrow$} & \textbf{cIoU $\uparrow$} &  \textbf{w/ Adap. $\uparrow$} & \textbf{AUC $\uparrow$} & \textbf{w/ Adap. $\uparrow$} \\ \midrule
    \textit{Prior Works:} & & & & & & & & \\
    Attention$_{\text{CVPR}18}$ & 18.50 & - & 30.20 & - & 66.00 & - & 55.80 & - \\
    CoarseToFine$_{\text{ECCV}20}$ & 29.10 & - & 34.80 & - & - & - & - & - \\
    LCBM$_{\text{WACV}22}$ & 32.20 & - & 36.60 & - & - & - & - & - \\
    LVS$_{\text{CVPR}21}$ & 34.40 & - & 38.20 & - & 71.90 & - & 58.20 & - \\
    HardPos$_{\text{ICASSP}22}$ & 34.60 & - & 38.00 & - & 76.80 & - & 59.20 & - \\
    SSPL$_{\text{CVPR}22}$ & 33.90 & - & 38.00 & - & 76.70 & - & 60.50 & - \\
    EZ-VSL$_{\text{ECCV}22}$ & 35.96 & 43.52 & 38.20 & 42.41 & 78.31 & 80.40 & 61.74 & 64.48 \\
    EZ-VSL (w/ OGL)$_{\text{ECCV}22}$ & 38.85 & 57.77 & 39.54 & 49.00 & 83.94 & 88.80 & 63.60 & 68.90 \\
    SSL-TIE$_{\text{ACM MM}22}$ & 38.63 & 51.92 & 39.65 & 48.06 & 79.50 & 84.80 & 61.20 & 65.64 \\
    SLAVC$_{\text{NeurIPS}22}$ & 37.79 & 49.41 & 39.40 & 45.79 & 83.60 & 85.20 & - & 66.70 \\
    SLAVC (w/ OGL)$_{\text{NeurIPS}22}$ & 39.80 & 59.08 & - & 49.73 & \textbf{86.00} & 90.00 & - & 69.28 \\
    MarginNCE$_{\text{ICASSP}23}$ & 38.25 & 50.76 & 39.06 & 46.39 & 83.94 & 88.80 & 63.20 & 68.92 \\
    MarginNCE (w/ OGL)$_{\text{ICASSP}23}$ & 39.78 & 59.29 & 40.01 & 50.23 & \underline{85.14} & \underline{91.60} & 64.55 & \underline{70.78} \\
    HearTheFlow$_{\text{WACV}23}$ & 39.40 & 54.56 & 40.00 & 48.01 & 84.80 & - & 64.00 & - \\
    HearTheFlow (w/ OGL)$_{\text{WACV}23}$ & 40.24 & 58.07 & 40.23 & 49.28 & 84.80 & - & 64.00 & - \\
    FNAC$_{\text{CVPR}23}$ & 39.50 & 47.00 & 39.66 & 43.30 & 84.73 & 89.20 & 63.76 & 69.78 \\
    FNAC (w/ OGL)$_{\text{CVPR}23}$ & 41.85 & 58.78 & 40.80 & 49.66 & 85.14 & \textbf{92.40} & 64.30 & 70.54 \\
    Alignment$_{\text{ICCV}23}$ & 41.42 & 57.25 & 40.76 & 49.32 & 83.20 & 88.00 & 64.00 & 69.36 \\
    Alignment (w/ OGL)$_{\text{ICCV}23}$ & 42.96 & 61.63 & 41.57 & 51.66 & 84.40 & \underline{91.60} & \textbf{65.14} & \textbf{71.70} \\
    \midrule
    \textit{Baselines:} & & & & & & & & \\
    \rowcolor{lightgray!25}
    \includegraphics[height=1em]{Figures/oracle.png}(\textit{\textbf{Oracle}}) CLIPSeg (w/ GT Text) & 49.50 &65.35 & 48.62 & 55.69 & - & - & - & - \\
    CLIPSeg (w/ WAV2CLIP Text) & 24.84 & 20.16 & 26.01 & 24.76 & 37.20 & 36.80 & 32.14 & 42.64 \\
    CLIPSeg (w/ CLAP Text) & 21.89 & 25.13 & 30.92 & 27.74 & 47.60 & 36.80 & 48.34 & 40.38 \\
    WAV2CLIP$_{\text{ICASSP}22}$ & 37.71 & - & 39.93 & - & 26.00 & - & 29.60 & - \\
    AudioCLIP$_{\text{ICASSP}22}$ & 44.15 & - & 46.23 & - & 47.20 & - & 45.22 & - \\
    \midrule
    \rowcolor{azure!10}
    \textit{\textbf{Ours}:} & & & & & & & & \\
    \rowcolor{azure!10}
    $\hookrightarrow$~w/ AV Alignment &49.46 & 64.02 & 46.32 & 54.14 & 80.80 & 86.80 & \underline{64.62} & 69.60 \\
    \rowcolor{azure!10}
    $\hookrightarrow$~w/ AV + LLM Alignment & \textbf{52.11} & \textbf{64.52} & \textbf{46.91} & \textbf{54.52} & 76.00 & 82.00 & 63.20 & 67.04 \\

    \toprule
    \end{tabular}}
    
    \caption{\textbf{Quantitative results on the VGG-SS and SoundNet-Flickr test sets}. All models are trained with 144K samples from VGG-Sound. SLAVC~\citep{slavc} does not provide AUC scores. SoundNet-Flickr has no GT text. Baselines and our method \emph{do not use OGL.} \includegraphics[height=1em]{Figures/oracle.png} method is Oracle.}
    \label{tab:quantitative_vgg_flickr}
\end{table*}
\subsection{Results} \label{ssec:quan}
In this section, we build on the evaluation protocol proposed by~\citep{senocak2024aligning}, which organizes sound source localization into three tasks: Single Sound Source Localization, Audio-Visual Segmentation, and Interactive Localization. We extend this by additionally including Audio-Visual Robustness and Multi-Source Localization, and present results for each.

\subsubsection{Single Sound Source Localization} \label{ssec:singlesource}
\newpara{Comparison on standard benchmarks.} This section presents a performance comparison between our method and existing approaches, including baselines. Experiments follow the standard evaluation protocol, based on previous works~\citep{chen2021localizing,ezvsl,sun2023learning,senocak2023alignment}. All models are trained on the VGGSound-144K dataset and evaluated on two widely used test sets: VGG-SS and SoundNet-Flickr. Notably, our approach does not utilize object-guided refinement (OGL). The quantitative results are reported in~\Tref{tab:quantitative_vgg_flickr}. 

There is a substantial performance gap between our method and existing self-supervised approaches on the VGG-SS, regardless of the variant used. Notably, our approach achieves the highest scores even compared to previous methods that utilize OGL, a post-processing technique designed to refine localization outputs. Although our method is trained purely in a self-supervised manner using an audio-visual correspondence objective -- similar to prior works -- the results clearly demonstrate that leveraging CLIP’s strong multimodal alignment significantly boosts performance. Furthermore, when LLM-based guidance is incorporated, our method gets additional gains. 
\begin{figure}[t]
    \centering
    \includegraphics[width=\linewidth]{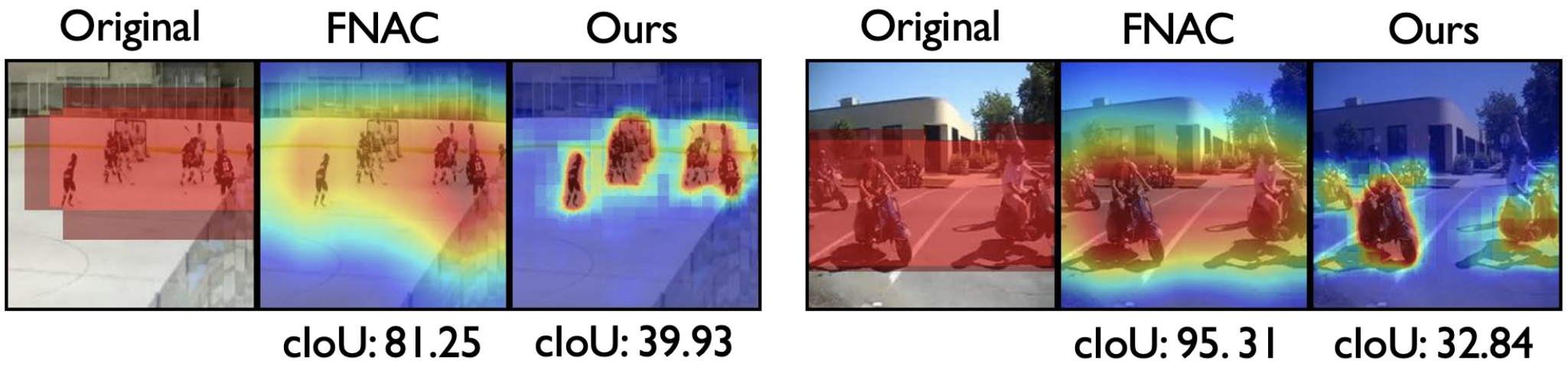}
    \caption{{\bf cIoU scores of SoundNet-Flickr samples.} }
    \label{fig:qualitative_cIoU}
\end{figure}
\begin{table}
\centering
\resizebox{0.99\linewidth}{!}{
\setlength{\tabcolsep}{3pt}
\begin{tabular}{clccccc}
\toprule
& \textbf{Method} & \textbf{cIoU} & \textbf{w/ Adap.} & \textbf{AUC} & \textbf{w/ Adap.} \\ 
    \midrule
    \multirow{14}{*}{\rotatebox[origin=c]{90}{\textbf{IS3}}}
    & LVS$_{\text{CVPR}21}$ & 33.4 & 39.4 & 39.0 & 41.1 \\
    & EZ-VSL$_{\text{ECCV}22}$ & 34.2 & 42.1 & 36.6 & 42.7 \\
    & SSL-TIE$_{\text{ACM MM}22}$ & 38.5 & 49.3 & 41.7 & 46.7 \\
    & SLAVC$_{\text{NeurIPS}22}$ & 36.9 & 45.0 & 40.2 & 42.7 \\
    & MarginNCE$_{\text{ICASSP}23}$ & 40.6 & 52.6 & 42.5 & 47.7 \\
    & FNAC$_{\text{CVPR}23}$ & 39.2 & 49.5 & 42.0 & 46.1 \\
    & Alignment$_{\text{ICCV}23}$ & 45.7 & 63.1 & 44.1 & 52.4 \\
    \cline{2-6}
    & \textit{Baselines:} & & & & \\
    & \cellcolor{lightgray!25}\includegraphics[height=1em]{Figures/oracle.png}(\textbf{\textit{Oracle}}) CLIPSeg (w/ GT Text) & \cellcolor{lightgray!25}74.3 & \cellcolor{lightgray!25}88.5 & \cellcolor{lightgray!25}59.3 & \cellcolor{lightgray!25}69.1 \\
    & CLIPSeg (w/ WAV2CLIP Text) & 26.7 & 55.4 & 25.8 & 49.8 \\
    & CLIPSeg (w/ CLAP Text) & 34.1 & 61.5 & 29.9 & 54.9 \\
    & WAV2CLIP$_{\text{ICASSP}22}$ & 40.5 & - & 43.9 & - \\
    & AudioCLIP$_{\text{ICASSP}22}$ & 39.7 & - & 44.2 & - \\
    \cline{2-6}
    & \cellcolor{azure!10}\textit{\textbf{Ours}:} &  \cellcolor{azure!10} & \cellcolor{azure!10} & \cellcolor{azure!10} & \cellcolor{azure!10} \\
    & \cellcolor{azure!10}$\hookrightarrow$~w/ AV Alignment & \cellcolor{azure!10} \underline{65.9} & \cellcolor{azure!10} \underline{79.0} & \cellcolor{azure!10} \underline{53.5} & \cellcolor{azure!10} \underline{62.9} \\
    & \cellcolor{azure!10}$\hookrightarrow$~w/ AV + LLM Alignment & \cellcolor{azure!10} \textbf{66.7} & \cellcolor{azure!10} \textbf{79.1} & \cellcolor{azure!10} \textbf{53.7} & \cellcolor{azure!10} \textbf{62.9} \\
    
\bottomrule
\bottomrule
    \multirow{16}{*}{\rotatebox[origin=c]{90}{\textbf{VPO-SS}}}
    & LVS$_{\text{CVPR}21}$ & 28.5 & 31.8 & 29.1 & 32.7 \\
    & EZ-VSL$_{\text{ECCV}22}$ & 26.6 & 30.3 & 29.0 & 32.9 \\
    & SSL-TIE$_{\text{ACM MM}22}$ & 31.7 & 39.1 & 30.6 & 36.7 \\
    & SLAVC$_{\text{NeurIPS}22}$ & 29.1 & 34.3 & 30.0 & 34.2 \\
    & MarginNCE$_{\text{ICASSP}23}$ & 32.1 & 35.6 & 30.3 & 35.1 \\
    & FNAC$_{\text{CVPR}23}$ & 31.5 & 36.3 & 30.8 & 34.8 \\
    & Alignment$_{\text{ICCV}23}$ & 30.4 & 38.7 & 30.4 & 36.2 \\
    \cline{2-6}
    & \textit{Baselines:} & & & & \\
    & \cellcolor{lightgray!25}\includegraphics[height=1em]{Figures/oracle.png}(\textit{\textbf{Oracle}}) CLIPSeg (w/ GT Text) & \cellcolor{lightgray!25} 60.9 & \cellcolor{lightgray!25} 74.2 & \cellcolor{lightgray!25} 49.0 & \cellcolor{lightgray!25} 59.8 \\
    & CLIPSeg (w/ WAV2CLIP Text) & 12.7 & 40.0 & 13.5 & 38.3 \\
    & CLIPSeg (w/ CLAP Text) & 35.4 & \underline{55.3} & 30.1 & \underline{49.1} \\
    & WAV2CLIP$_{\text{ICASSP}22}$ & 26.9 & - & 28.5 & - \\
    & AudioCLIP$_{\text{ICASSP}22}$ & \underline{43.5} & - & \textbf{44.2} & - \\
    \cline{2-6}
    & \cellcolor{azure!10}\textit{\textbf{Ours}:} &  \cellcolor{azure!10} & \cellcolor{azure!10} & \cellcolor{azure!10} & \cellcolor{azure!10} \\
    & \cellcolor{azure!10} $\hookrightarrow$~w/ AV Alignment & \cellcolor{azure!10} 38.1 & \cellcolor{azure!10} 52.3 & \cellcolor{azure!10} 33.9 & \cellcolor{azure!10} 44.9 \\
    & \cellcolor{azure!10} $\hookrightarrow$~w/ AV +  LLM Alignment  & \cellcolor{azure!10} \textbf{48.3} & \cellcolor{azure!10} \textbf{63.6} & \cellcolor{azure!10} \underline{40.4} & \cellcolor{azure!10} \textbf{51.5} \\

\bottomrule
\bottomrule
    \multirow{16}{*}{\rotatebox[origin=c]{90}{\textbf{VPO-MS}}}
    & LVS$_{\text{CVPR}21}$ & 25.0 & 28.9 & 27.8 & 30.9 \\
    & EZ-VSL$_{\text{ECCV}22}$ & 25.4 & 31.0 & 28.7 & 32.6 \\
    & SSL-TIE$_{\text{ACM MM}22}$ & 27.5 & 35.4 & 29.4 & 35.1 \\
    & SLAVC$_{\text{NeurIPS}22}$ & 27.1 & 33.9 & 29.2 & 34.0 \\
    & MarginNCE$_{\text{ICASSP}23}$ & 28.7 & 33.5 & 29.5 & 34.1 \\
    & FNAC$_{\text{CVPR}23}$ & 28.4 & 34.9 & 29.8 & 34.2 \\
    & Alignment$_{\text{ICCV}23}$ & 29.0 & 36.4 & 29.7 & 35.2 \\
    \cline{2-6}
    & \textit{Baselines:} & & & & \\
    & \cellcolor{lightgray!25}\includegraphics[height=1em]{Figures/oracle.png}(\textit{\textbf{Oracle}}) CLIPSeg (w/ GT Text) & \cellcolor{lightgray!25} 54.7 & \cellcolor{lightgray!25} 70.6 & \cellcolor{lightgray!25} 45.6 & \cellcolor{lightgray!25} 58.0 \\
    & CLIPSeg (w/ WAV2CLIP Text) & 13.4 & 40.1 & 14.0 & 38.1 \\
    & CLIPSeg (w/ CLAP Text) & 31.3 & \underline{51.5} & 27.9 & \underline{45.7} \\
    & WAV2CLIP$_{\text{ICASSP}22}$ & 27.4 & - & 29.0 & - \\
    & AudioCLIP$_{\text{ICASSP}22}$ & 37.7 & - & \underline{39.0} & - \\
    \cline{2-6}
    & \cellcolor{azure!10}\textit{\textbf{Ours}:} &  \cellcolor{azure!10} & \cellcolor{azure!10} & \cellcolor{azure!10} & \cellcolor{azure!10} \\
    & \cellcolor{azure!10} $\hookrightarrow$~w/ AV Alignment & \cellcolor{azure!10} \underline{38.2} & \cellcolor{azure!10} 51.3 & \cellcolor{azure!10} 34.8 & \cellcolor{azure!10} 45.2 \\
    & \cellcolor{azure!10} $\hookrightarrow$~w/ AV +  LLM Alignment  & \cellcolor{azure!10} \textbf{45.0} & \cellcolor{azure!10} \textbf{59.1} & \cellcolor{azure!10} \textbf{39.6} & \cellcolor{azure!10} \textbf{49.6} \\

\bottomrule
\bottomrule
    \multirow{16}{*}{\rotatebox[origin=c]{90}{\textbf{AVS-Bench S4}}}
    & LVS$_{\text{CVPR}21}$ & 42.0 & 51.2 & 41.0 & 47.2 \\
    & EZ-VSL$_{\text{ECCV}22}$ & 44.9 & 52.4 & 41.9 & 47.5 \\
    & SSL-TIE$_{\text{ACM MM}22}$ & 47.4 & 60.8 & 43.2 & 53.3 \\
    & SLAVC$_{\text{NeurIPS}22}$ & 46.8 & 58.2 & 43.2 & 50.7 \\
    & MarginNCE$_{\text{ICASSP}23}$ & 47.7 & 59.0 & 43.7 & 51.8 \\
    & FNAC$_{\text{CVPR}23}$ & 48.4 & 58.5 & 43.8 & 50.7 \\
    & Alignment$_{\text{ICCV}23}$ & 52.1 & 67.4 & 45.0 & 55.9 \\
    \cline{2-6}
    & \textit{Baselines:} & & & & \\
    & \cellcolor{lightgray!25}\includegraphics[height=1em]{Figures/oracle.png}(\textit{\textbf{Oracle}}) CLIPSeg (w/ GT Text) & \cellcolor{lightgray!25}66.5 & \cellcolor{lightgray!25}80.2 & \cellcolor{lightgray!25}54.1 & \cellcolor{lightgray!25}64.9 \\
    & CLIPSeg (w/ WAV2CLIP Text) & 36.3 & 62.6 & 31.9 & \underline{64.9} \\
    & CLIPSeg (w/ CLAP Text) & 56.4 & 74.0 & 47.2 & 61.5 \\
    & WAV2CLIP$_{\text{ICASSP}22}$ & 43.2 & - & 45.1 & - \\
    & AudioCLIP$_{\text{ICASSP}22}$ & 60.6 & - & \textbf{61.1} & - \\
    \cline{2-6}
    & \cellcolor{azure!10}\textit{\textbf{Ours}:} &  \cellcolor{azure!10} & \cellcolor{azure!10} & \cellcolor{azure!10} & \cellcolor{azure!10} \\
    & \cellcolor{azure!10} $\hookrightarrow$~w/ AV Alignment & \cellcolor{azure!10} 63.5 & \cellcolor{azure!10} \textbf{81.2} & \cellcolor{azure!10} 54.9 & \cellcolor{azure!10} \underline{64.9} \\
    & \cellcolor{azure!10} $\hookrightarrow$~w/ AV +  LLM Alignment  & \cellcolor{azure!10} \textbf{68.2} & \cellcolor{azure!10} \underline{80.9} & \cellcolor{azure!10} \underline{55.7} & \cellcolor{azure!10} \textbf{65.3} \\

\bottomrule
\bottomrule
    \multirow{16}{*}{\rotatebox[origin=c]{90}{\textbf{ADE20K}}}
    & LVS$_{\text{CVPR}21}$ & 33.0 & 36.8 & 35.0 & 39.2 \\
    & EZ-VSL$_{\text{ECCV}22}$ & 35.8 & 45.3 & 36.4 & 42.5 \\
    & SSL-TIE$_{\text{ACM MM}22}$ & 38.7 & 48.1 & 37.9 & 44.9 \\
    & SLAVC$_{\text{NeurIPS}22}$ & 38.7 & 45.3 & 38.5 & 45.7 \\
    & MarginNCE$_{\text{ICASSP}23}$ & 34.9 & 44.3 & 37.4 & 44.8 \\
    & FNAC$_{\text{CVPR}23}$ & 38.7 & 42.5 & 37.8 & 43.5 \\
    & Alignment$_{\text{ICCV}23}$ & 40.6 & 51.9 & 38.5 & 47.4 \\
    \cline{2-6}
    & \textit{Baselines:} & & & & \\
    & \cellcolor{lightgray!25}\includegraphics[height=1em]{Figures/oracle.png} (\textit{\textbf{Oracle}}) CLIPSeg (w/ GT Text) & \cellcolor{lightgray!25}- & \cellcolor{lightgray!25}- & \cellcolor{lightgray!25}- & \cellcolor{lightgray!25}- \\
    & CLIPSeg (w/ WAV2CLIP Text) & 19.8 & 37.7 & 18.3 & 35.3 \\
    & CLIPSeg (w/ CLAP Text) & 22.6 & 40.6 & 21.4 & 39.1 \\
    & WAV2CLIP$_{\text{ICASSP}22}$ & 18.9 & - & 24.4 & - \\
    & AudioCLIP$_{\text{ICASSP}22}$ & 26.4 & - & 32.7 & - \\
    \cline{2-6}
    & \cellcolor{azure!10}\textit{\textbf{Ours}:} &  \cellcolor{azure!10} & \cellcolor{azure!10} & \cellcolor{azure!10} & \cellcolor{azure!10} \\
    & \cellcolor{azure!10} $\hookrightarrow$~w/ AV Alignment & \cellcolor{azure!10} \underline{45.3} & \cellcolor{azure!10} \underline{58.5} & \cellcolor{azure!10} \underline{42.4} & \cellcolor{azure!10} \underline{54.6} \\
    & \cellcolor{azure!10} $\hookrightarrow$~w/ AV +  LLM Alignment  & \cellcolor{azure!10} \textbf{53.8} & \cellcolor{azure!10} \textbf{67.0} & \cellcolor{azure!10} \textbf{48.2} & \cellcolor{azure!10} \textbf{58.6} \\

\bottomrule
    
    \end{tabular}}
\caption{\textbf{Sound source localization results on recent benchmarks.} All models are trained on the VGGSound-144K dataset. Results without object guided refinement
(OGL) are reported.}
    {
    \label{tab:bbox_combined}}

\end{table}

Interestingly, we observe that the zero-shot performance of our model on the SoundNet-Flickr test set lags behind existing models. We hypothesize this stems from the fact that our model generates more fine-grained outputs, resembling segmentation. However, the ground-truth bounding boxes for this test set are relatively coarse, causing our method to yield lower cIoU scores even when it successfully highlights the sounding region. As shown in~\Fref{fig:qualitative_cIoU}, our model clearly identifies the sounding regions, yet these results still produce lower cIoU scores. This outcome is consistent with the quantitative results in~\Tref{tab:quantitative_vgg_flickr}, which demonstrate that our method on SoundNet-Flickr lags behind other methods due to the fact that the GT boxes and the localization results of competing methods are coarse.

Next, we compare our method against the strong baselines introduced earlier and observe that it consistently outperforms them. Notably, our method does not explicitly utilize text input to highlight object regions via CLIPSeg, as these baselines do. This suggests that our audio-visual correspondence objective is effective in learning robust audio-visual relationships, enabling the AudioTokenizer and Audio-Driven Embedder to accurately project the true audio context into meaningful embeddings. Interestingly, our AV Alignment variant achieves performance on par with the CLIPSeg (w/ GT Text) baseline on VGG-SS, which serves as an oracle. This baseline represents a text-conditioned, open-world segmentation approach that leverages the ground-truth class labels of the test samples. This comparison highlights the importance of incorporating audio context in a descriptive and semantically rich manner. Furthermore, the performance gap between CLIPSeg (w/ GT Text) and its variants using WAV2CLIP or CLAP shows that the zero-shot performance of CLIPSeg is highly dependent on the quality of the text input. This is likely because the text retrieved from WAV2CLIP or CLAP tends to be noisier than the ground-truth text. Nevertheless, it is important to emphasize that sound source localization is an unlabeled, audio-input-driven task, where ground-truth class labels are often unavailable across datasets. As a result, methods that rely on class label text-inputs cannot be applied universally (only in the datasets with GT class labels). These class label–conditioned methods serve primarily as oracle baselines to illustrate the robustness and effectiveness of our model.

Finally, we compare our model with AudioCLIP and WAV2CLIP, both of which are trained contrastively on image-audio pairs using the pre-trained CLIP model. As shown in~\Tref{tab:quantitative_vgg_flickr}, our method outperforms both approaches. This suggests that our Audio-Driven Embedder, trained with the audio-visual alignment objective, is more effective at learning robust audio-visual correspondence than these prior methods, despite all leveraging pre-trained CLIP knowledge.

\begin{figure*}[t]
    \centering
    \includegraphics[width=\linewidth]{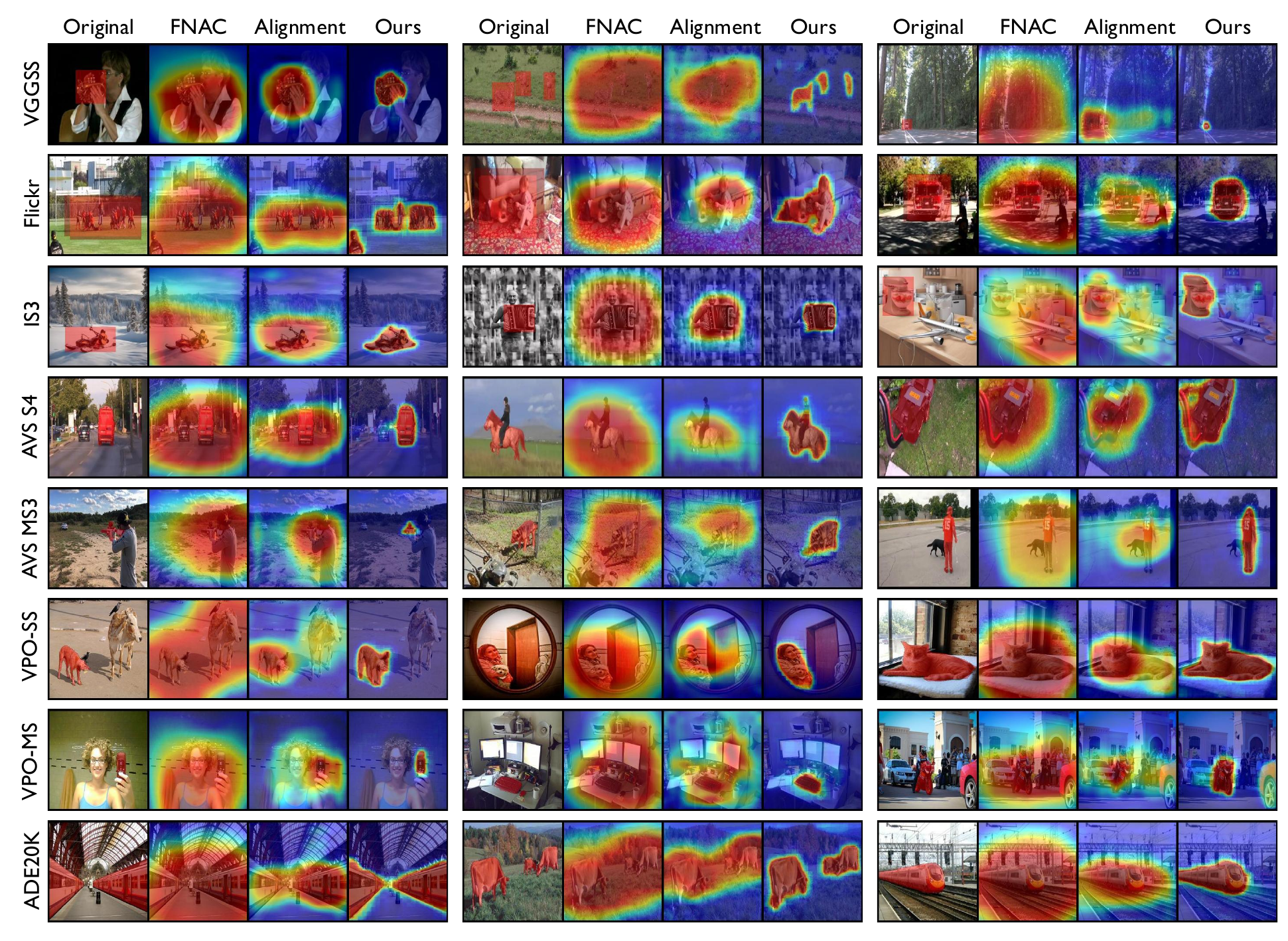}
    \caption{{\bf 
    Qualitative sound source localization results on various datasets.
     }}
    \label{fig:qualitatives}
\end{figure*}

\newpara{Comparison on recent benchmarks.} We also compare our method on recent benchmarks, following the evaluation protocol proposed by~\citep{senocak2024aligning}. These benchmarks cover a diverse range of characteristics -- for example, some datasets contain real-world samples, while others consist of synthetic data (\eg, IS3), and they span distinct semantic categories. While these benchmarks provide segmentation maps for the sounding objects, we follow the conventional detection-based evaluation protocol for the single sound source localization task. Specifically, we convert the segmentation maps into bounding boxes and compute cIoU, AUC, and their adaptive variants, following the evaluation procedure described in~\citep{senocak2024aligning}. The results are presented in~\Tref{tab:bbox_combined}.

Similar to the previous comparison, our method demonstrates superior performance across all five benchmarks when compared to prior works and baselines (excluding the Oracle method). On some datasets, our \textit{AV + LLM Alignment} variant even surpasses the Oracle. The consistent high performance of our method across all datasets indicates the generalizability of our approach. We also observe that incorporating additional LLM guidance during training brings significant improvements over the AV-only alignment model on most datasets (\eg, +10.2 cIoU and +11.3 in cIoU Adaptive in VPO-SS). These findings further emphasize the state-of-the-art performance of our proposed method to date.

\newpara{Qualitative Results.} ~\Fref{fig:qualitatives} shows the qualitative comparison between our method and recent prior works on various datasets. The visualized samples illustrate that the localized regions from our proposed method are more compact and fine-grained compared to the other methods. For example, regardless of the test set, our model can accurately localize small-sized sounding objects compared to recent methods. Moreover, our model accurately highlights multiple sound sources and separates them, while other methods tend to cover the entire area as one large region (last row, columns 1–2).



\begin{table}[t]
    \centering
    \resizebox{1.0\linewidth}{!}{
    \setlength{\tabcolsep}{3pt}
    \begin{tabular}{lcccccc}
    \toprule
     & \multicolumn{3}{c}{\textbf{Extended VGG-SS}} & \multicolumn{3}{c}{\textbf{Extended Flickr}} \\
    Method & \textbf{AP} & \textbf{max-F1} & \textbf{LocAcc} & \textbf{AP} & \textbf{max-F1} & \textbf{LocAcc} \\ \midrule
    SLAVC$_{\text{NeurIPS}22}$ & 32.95 & 40.00 & 37.79 & 51.63 & 59.10 & 83.60 \\
    MarginNCE$_{\text{ICASSP}23}$ & 30.58 & 36.80 & 38.25 & 57.99 & 61.80 & \underline{83.94}\\
    FNAC$_{\text{CVPR}23}$ & 23.48 & 33.70 & 39.50 & 50.40 & 62.30 & \textbf{84.73} \\
    Alignment$_{\text{ICCV}23}$ & 34.73 & 40.70 & 39.94 & 64.43 & 66.90 & 79.60 \\
    \midrule
    \textit{Baselines:} \\
    WAV2CLIP$_{\text{ICASSP}22}$ & 26.67 & 33.00 & 37.71 & 20.99 & 24.80 & 29.60 \\
    AudioCLIP$_{\text{ICASSP}22}$ & 23.79 & 32.80 & 44.15 & 34.00 & 38.80 & 45.22 \\
    \midrule
    \rowcolor{azure!10}
    \textit{\textbf{Ours}:} & & & & & & \\
    \rowcolor{azure!10}
    $\hookrightarrow$~w/ AV Align. & \underline{40.79} & \underline{49.10} & \underline{49.46} & \textbf{76.07} & \textbf{73.20} & 80.80 \\
    \rowcolor{azure!10}
    $\hookrightarrow$~w/ AV + LLM Align. & \textbf{46.72} & \textbf{51.90} & \textbf{52.11} & \underline{75.21} & \underline{72.40} & 76.00 \\    
    \toprule
    \end{tabular}}
    
    \caption{\textbf{Quantitative results on Extended VGG-SS and Extended Flickr-SoundNet benchmark.} All models are trained with 144K samples from VGG-Sound.}
    \label{tab:extended}
\end{table}

\begin{table}
\centering
\resizebox{1.0\linewidth}{!}{
\setlength{\tabcolsep}{3pt}
\begin{tabular}{l|ccccc|c}
\toprule
 & \multicolumn{6}{c}{\textbf{Extended Flickr}} \\
Method & AP $\uparrow$ & max-F1 $\uparrow$ & Precision $\uparrow$ & Recall $\uparrow$ & LocAcc $\uparrow$ & Recall - Precision $\downarrow$\\
\midrule
SLAVC$_{\text{NeurIPS}22}$ & 51.63 & 59.10 & 43.78 & \textbf{90.78} & \underline{83.60} & 47.00 \\
FNAC$_{\text{CVPR}23}$ & 50.40 & 62.3 & 47.00 & \underline{89.52} & \textbf{84.73} & 42.52 \\
\midrule
\rowcolor{azure!10}
\textbf{Ours:} & & & & & & \\
\rowcolor{azure!10}
$\hookrightarrow$~w/ AV Alignment & \textbf{76.07} & \textbf{73.20} & \textbf{63.67} & 86.04 & 80.80 & \textbf{22.37} \\
\rowcolor{azure!10}
$\hookrightarrow$~w/ AV + LLM Alignment & \underline{75.21} & \underline{72.40} & \underline{61.33} & 88.46 & 76.00 & \underline{27.13} \\
\bottomrule
\end{tabular}}
\caption{\textbf{Analysis on Extended SoundNet-Flickr dataset.}}
\label{tab:ext_analysis}
\end{table}

\begin{figure}[t]
    \centering
    \includegraphics[width=\linewidth]{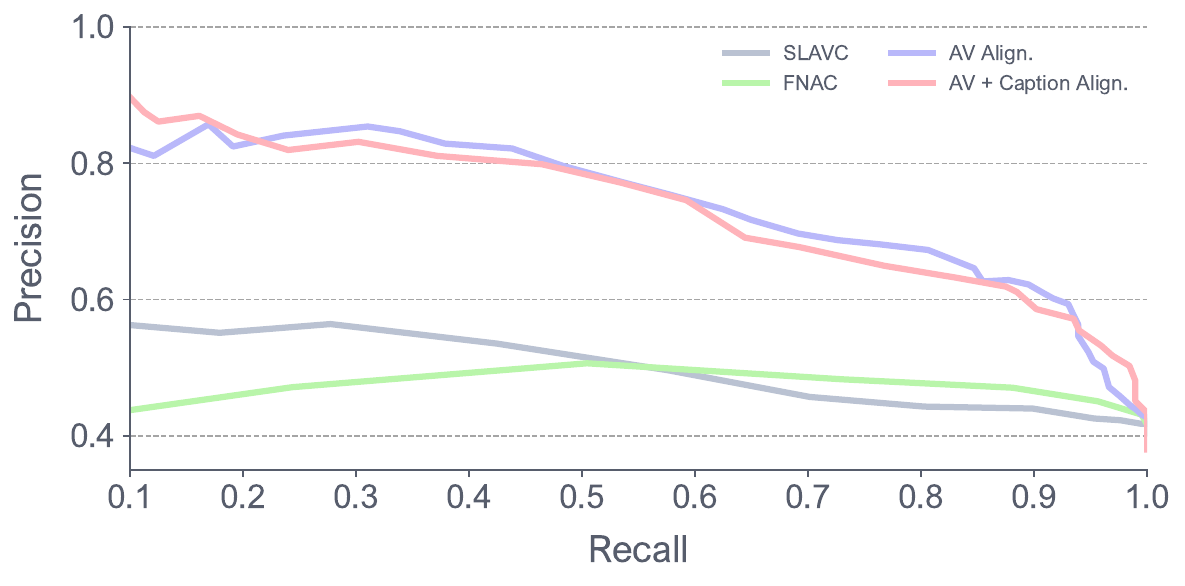}
    \caption{{\bf Precision-Recall Curve on Extended Flickr-SoundNet benchmark.}}
    \label{fig:pr_analysis}
\end{figure}

\subsubsection{Audio-Visual Robustness} \label{ssec:robustness}
Existing benchmarks typically consist of sounding objects/regions in the scene. However, in reality, silent objects or off-screen audio are also common occurrences. Mo et al.~\citep{slavc} propose a new evaluation that extends the existing benchmarks to include non-audible frames, non-visible sound sources, and mismatched audio-visual pairs. In this evaluation scenario, it is expected that sound localization methods should not highlight an object/region if the audio and visual signals are mismatched. The experiments conducted using the Extended Flickr-SoundNet/VGG-SS datasets in~\Tref{tab:extended} demonstrate that our method outperforms all the existing methods and baselines. The superiority of our method indicates that it learns a strong alignment of audio and visual embeddings with the help of our AudioTokenizer and leveraging CLIP alignment knowledge as this task requires a robust semantic relationship between the cross-modalities. One interesting observation is that, even though baseline approaches leverage CLIP, their performance is lower than ours due to the absence of audio-visual alignment supervision.
\begin{table}
\centering
\resizebox{0.96\linewidth}{!}{
\begin{tabular}{clccccc}
\toprule
& \textbf{Method} & \textbf{mIoU} & \textbf{w/ Adap.} & \textbf{F-Score} & \textbf{w/ Adap.} \\ 
    \midrule
    \multirow{14}{*}{\rotatebox[origin=c]{90}{\textbf{IS3}}}
    & LVS$_{\text{CVPR}21}$ & 23.8 & 23.8 & 29.7 & 34.9 \\
    & EZ-VSL$_{\text{ECCV}22}$ & 24.5 & 26.3 & 30.3 & 37.7 \\
    & SSL-TIE$_{\text{ACM MM}22}$ & 26.0 & 32.6 & 32.1 & 45.4 \\
    & SLAVC$_{\text{NeurIPS}22}$ & 24.3 & 26.7 & 30.1 & 37.9 \\
    & MarginNCE$_{\text{ICASSP}23}$ & 26.1 & 30.4 & 31.9 & 42.9 \\
    & FNAC$_{\text{CVPR}23}$ & 25.3 & 27.4 & 31.1 & 39.1 \\
    & Alignment$_{\text{ICCV}23}$ & 27.3 & 35.7 & 33.1 & 49.0 \\
    \cline{2-6}
    & \textit{Baselines:} & & & & \\
    & \cellcolor{lightgray!25}\includegraphics[height=1em]{Figures/oracle.png}(\textit{\textbf{Oracle}}) CLIPSeg (w/ GT Text) & \cellcolor{lightgray!25}58.4 & \cellcolor{lightgray!25}74.2 & \cellcolor{lightgray!25}65.1 & \cellcolor{lightgray!25}82.9 \\
    & CLIPSeg (w/ WAV2CLIP Text) & 18.5 & 41.9 & 22.2 & 51.9 \\
    & CLIPSeg (w/ CLAP Text) & 28.8 & 54.7 & 33.9 & 64.7 \\
    & WAV2CLIP$_{\text{ICASSP}22}$ & 25.6 & - & 31.7 & - \\
    & AudioCLIP$_{\text{ICASSP}22}$ & 23.7 & - & 28.6 & - \\
    \cline{2-6}
    & \cellcolor{azure!10}\textit{\textbf{Ours}:} &  \cellcolor{azure!10} & \cellcolor{azure!10} & \cellcolor{azure!10} & \cellcolor{azure!10} \\
    & \cellcolor{azure!10}$\hookrightarrow$~w/ AV Alignment & \cellcolor{azure!10} \underline{55.6} & \cellcolor{azure!10} \underline{65.9} & \cellcolor{azure!10} \underline{62.9} & \cellcolor{azure!10} \textbf{74.1} \\
    & \cellcolor{azure!10}$\hookrightarrow$~w/ AV + LLM Alignment & \cellcolor{azure!10} \textbf{56.3} & \cellcolor{azure!10} \textbf{66.0} & \cellcolor{azure!10} \textbf{63.3} & \cellcolor{azure!10} \underline{73.8} \\
    
\bottomrule
\bottomrule
    \multirow{14}{*}{\rotatebox[origin=c]{90}{\textbf{VPO-SS}}}
    & LVS$_{\text{CVPR}21}$ & 20.3 & 21.4 & 25.5 & 29.7 \\
    & EZ-VSL$_{\text{ECCV}22}$ & 20.0 & 26.3 & 25.3 & 37.7 \\
    & SSL-TIE$_{\text{ACM MM}22}$ & 21.0 & 26.5 & 26.4 & 36.2 \\
    & SLAVC$_{\text{NeurIPS}22}$ & 20.6 & 21.7 & 25.8 & 30.2 \\
    & MarginNCE$_{\text{ICASSP}23}$ & 20.8 & 24.6 & 26.1 & 33.9 \\
    & FNAC$_{\text{CVPR}23}$ & 21.1 & 22.5 & 26.3 & 31.2 \\
    & Alignment$_{\text{ICCV}23}$ & 21.0 & 23.6 & 26.3 & 32.8 \\
    \cline{2-6}
    & \textit{Baselines:} & & & & \\
    & \cellcolor{lightgray!25}\includegraphics[height=1em]{Figures/oracle.png}(\textit{\textbf{Oracle}}) CLIPSeg (w/ GT Text)  & \cellcolor{lightgray!25}51.6 & \cellcolor{lightgray!25}62.4 & \cellcolor{lightgray!25}57.9 & \cellcolor{lightgray!25}69.9 \\
    & CLIPSeg (w/ WAV2CLIP Text) & 9.6 & 30.8 & 18.4 & 38.9 \\
    & CLIPSeg (w/ CLAP Text) & 28.8 & \underline{48.5} & 32.2 & \underline{55.9} \\
    & WAV2CLIP$_{\text{ICASSP}22}$ & 17.7 & - & 22.3 & - \\
    & AudioCLIP$_{\text{ICASSP}22}$ & 26.1 & - & 30.5 & - \\
    \cline{2-6}
    & \cellcolor{azure!10}\textit{\textbf{Ours}:} &  \cellcolor{azure!10} & \cellcolor{azure!10} & \cellcolor{azure!10} & \cellcolor{azure!10} \\
    & \cellcolor{azure!10}$\hookrightarrow$~w/ AV Alignment & \cellcolor{azure!10} \underline{32.8} & \cellcolor{azure!10} 42.9 & \cellcolor{azure!10} \underline{37.7} & \cellcolor{azure!10} 49.9 \\
    & \cellcolor{azure!10}$\hookrightarrow$~w/ AV + LLM Alignment & \cellcolor{azure!10} \textbf{39.5} & \cellcolor{azure!10} \textbf{51.7} & \cellcolor{azure!10} \textbf{44.7} & \cellcolor{azure!10} \textbf{58.4} \\

\bottomrule
\bottomrule
    \multirow{14}{*}{\rotatebox[origin=c]{90}{\textbf{VPO-MS}}}
    & LVS$_{\text{CVPR}21}$ & 17.8 & 18.2 & 22.7 & 25.6 \\
    & EZ-VSL$_{\text{ECCV}22}$ & 18.5 & 19.4 & 23.4 & 27.4 \\
    & SSL-TIE$_{\text{ACM MM}22}$ & 19.1 & 24.2 & 24.0 & 32.7 \\
    & SLAVC$_{\text{NeurIPS}22}$ & 18.7 & 20.1 & 23.6 & 28.0 \\
    & MarginNCE$_{\text{ICASSP}23}$ & 19.2 & 22.4 & 24.1 & 30.9 \\
    & FNAC$_{\text{CVPR}23}$ & 19.1 & 20.7 & 24.0 & 28.7 \\
    & Alignment$_{\text{ICCV}23}$ & 19.7 & 23.1 & 24.6 & 31.7 \\
    \cline{2-6}
    & \textit{Baselines:} & & & & \\
    & \cellcolor{lightgray!25}\includegraphics[height=1em]{Figures/oracle.png}(\textit{\textbf{Oracle}}) CLIPSeg (w/ GT Text)  & \cellcolor{lightgray!25}46.3 & \cellcolor{lightgray!25}58.6 & \cellcolor{lightgray!25}53.4 & \cellcolor{lightgray!25}66.9 \\
    & CLIPSeg (w/ WAV2CLIP Text) & 11.0 & 30.2 & 16.9 & 37.8 \\
    & CLIPSeg (w/ CLAP Text) & 26.5 & \underline{45.1} & 30.6 & \underline{52.8} \\
    & WAV2CLIP$_{\text{ICASSP}22}$ & 18.9 & - & 24.1 & - \\
    & AudioCLIP$_{\text{ICASSP}22}$ & 24.6 & - & 29.0 & - \\
    \cline{2-6}
    & \cellcolor{azure!10}\textit{\textbf{Ours}:} &  \cellcolor{azure!10} & \cellcolor{azure!10} & \cellcolor{azure!10} & \cellcolor{azure!10} \\
    & \cellcolor{azure!10}$\hookrightarrow$~w/ AV Alignment & \cellcolor{azure!10} \underline{34.3} & \cellcolor{azure!10} 43.2 & \cellcolor{azure!10} \underline{40.0} & \cellcolor{azure!10} 50.1 \\
    & \cellcolor{azure!10}$\hookrightarrow$~w/ AV + LLM Alignment & \cellcolor{azure!10} \textbf{39.5} & \cellcolor{azure!10} \textbf{49.3} & \cellcolor{azure!10} \textbf{45.5} & \cellcolor{azure!10} \textbf{56.2} \\

\bottomrule
\bottomrule
    \multirow{14}{*}{\rotatebox[origin=c]{90}{\textbf{AVS-Bench S4}}}
    & LVS$_{\text{CVPR}21}$ & 27.0 & 30.5 & 33.4 & 42.4 \\
    & EZ-VSL$_{\text{ECCV}22}$ & 27.7 & 30.7 & 34.1 & 42.8 \\
    & SSL-TIE$_{\text{ACM MM}22}$ & 28.9 & 38.9 & 352. & 52.5 \\
    & SLAVC$_{\text{NeurIPS}22}$ & 28.0 & 32.8 & 34.4 & 45.5 \\
    & MarginNCE$_{\text{ICASSP}23}$ & 28.9 & 35.4 & 35.3 & 48.6 \\
    & FNAC$_{\text{CVPR}23}$ & 28.8 & 33.0 & 35.3 & 45.6 \\
    & Alignment$_{\text{ICCV}23}$ & 30.1 & 40.6 & 36.3 & 54.3 \\
    \cline{2-6}
    & \textit{Baselines:} & & & & \\
    & \cellcolor{lightgray!25}\includegraphics[height=1em]{Figures/oracle.png}(\textit{\textbf{Oracle}}) CLIPSeg (w/ GT Text)  & \cellcolor{lightgray!25}51.3 & \cellcolor{lightgray!25}63.5 & \cellcolor{lightgray!25}58.0 & \cellcolor{lightgray!25}72.8 \\
    & CLIPSeg (w/ WAV2CLIP Text) & 26.5 & 46.7 & 30.6 & 57.4 \\
    & CLIPSeg (w/ CLAP Text) & 41.3 & 58.5 & 46.8 & 67.9 \\
    & WAV2CLIP$_{\text{ICASSP}22}$ & 28.7 & - & 35.4 & - \\
    & AudioCLIP$_{\text{ICASSP}22}$ & 36.6 & - & 42.2 & - \\
    \cline{2-6}
    & \cellcolor{azure!10}\textit{\textbf{Ours}:} &  \cellcolor{azure!10} & \cellcolor{azure!10} & \cellcolor{azure!10} & \cellcolor{azure!10} \\
    & \cellcolor{azure!10}$\hookrightarrow$~w/ AV Alignment & \cellcolor{azure!10} \underline{59.8} & \cellcolor{azure!10} \underline{66.8} & \cellcolor{azure!10} \underline{69.0} & \cellcolor{azure!10} \textbf{75.9} \\
    & \cellcolor{azure!10}$\hookrightarrow$~w/ AV + LLM Alignment & \cellcolor{azure!10} \textbf{61.8} & \cellcolor{azure!10} \textbf{67.1} & \cellcolor{azure!10} \textbf{69.6} & \cellcolor{azure!10} \underline{75.2} \\

\bottomrule
\bottomrule

    \multirow{14}{*}{\rotatebox[origin=c]{90}{\textbf{AVS-Bench MS3}}}
    & LVS$_{\text{CVPR}21}$ & 22.8 & 26.8 & 25.1 & 28.9 \\
    & EZ-VSL$_{\text{ECCV}22}$ & 22.6 & 27.8 & 25.0 & 30.9 \\
    & SSL-TIE$_{\text{ACM MM}22}$ & 23.5 & 32.7 & 25.9 & 37.8 \\
    & SLAVC$_{\text{NeurIPS}22}$ & 22.1 & 26.1 & 24.3 & 28.5 \\
    & MarginNCE$_{\text{ICASSP}23}$ & 23.1 & 30.1 & 25.5 & 35.4 \\
    & FNAC$_{\text{CVPR}23}$ & 23.2 & 30.4 & 25.5 & 34.2 \\
    & Alignment$_{\text{ICCV}23}$ & 23.7 & 31.4 & 26.2 & 35.9 \\
    \cline{2-6}
    & \textit{Baselines:} & & & & \\
    & \cellcolor{lightgray!25}\includegraphics[height=1em]{Figures/oracle.png}(\textit{\textbf{Oracle}}) CLIPSeg (w/ GT Text) & \cellcolor{lightgray!25}50.9 & \cellcolor{lightgray!25}55.3 & \cellcolor{lightgray!25}55.9 & \cellcolor{lightgray!25}66.2 \\
    & CLIPSeg (w/ WAV2CLIP Text) & 30.8 & 39.8 & 30.0 & 49.5 \\
    & CLIPSeg (w/ CLAP Text) & \underline{43.0} & \underline{46.5} & 45.4 & \underline{56.3} \\
    & WAV2CLIP$_{\text{ICASSP}22}$ & 25.1 & - & 23.8 & - \\
    & AudioCLIP$_{\text{ICASSP}22}$ & 27.1 & - & 26.5 & - \\
    \cline{2-6}
    & \cellcolor{azure!10}\textit{\textbf{Ours}:} &  \cellcolor{azure!10} & \cellcolor{azure!10} & \cellcolor{azure!10} & \cellcolor{azure!10} \\
    & \cellcolor{azure!10}$\hookrightarrow$~w/ AV Alignment & \cellcolor{azure!10} 41.1 & \cellcolor{azure!10} 44.4 & \cellcolor{azure!10} \underline{46.7} & \cellcolor{azure!10} 54.2 \\
    & \cellcolor{azure!10}$\hookrightarrow$~w/ AV + LLM Alignment & \cellcolor{azure!10} \textbf{43.6} & \cellcolor{azure!10} \textbf{47.4} & \cellcolor{azure!10} \textbf{48.4} & \cellcolor{azure!10} \textbf{56.9} \\

\bottomrule
\bottomrule

    \multirow{14}{*}{\rotatebox[origin=c]{90}{\textbf{ADE20K}}}
    & LVS$_{\text{CVPR}21}$ & 22.1 & 20.0 & 27.2 & 29.5 \\
    & EZ-VSL$_{\text{ECCV}22}$ & 22.3 & 20.4 & 27.3 & 29.8 \\
    & SSL-TIE$_{\text{ACM MM}22}$ & 23.6 & 26.2 & 28.7 & 37.3 \\
    & SLAVC$_{\text{NeurIPS}22}$ & 24.2 & 22.0 & 29.0 & 32.4 \\
    & MarginNCE$_{\text{ICASSP}23}$ & 24.1 & 22.3 & 28.9 & 33.8 \\
    & FNAC$_{\text{CVPR}23}$ & 23.9 & 21.8 & 28.7 & 32.1 \\
    & Alignment$_{\text{ICCV}23}$ & 25.0 & 26.9 & 30.0 & 38.3 \\
    \cline{2-6}
    & \textit{Baselines:} & & & & \\
    & \cellcolor{lightgray!25}\includegraphics[height=1em]{Figures/oracle.png}(\textit{\textbf{Oracle}}) CLIPSeg (w/ GT Text) & \cellcolor{lightgray!25}- & \cellcolor{lightgray!25}- & \cellcolor{lightgray!25}- & \cellcolor{lightgray!25}- \\
    & CLIPSeg (w/ WAV2CLIP Text) & 16.6 & 34.9 & 20.8 & 43.3 \\
    & CLIPSeg (w/ CLAP Text) & 20.0 & 38.4 & 24.0 & 48.0 \\
    & WAV2CLIP$_{\text{ICASSP}22}$ & 16.9 & - & 23.3 & - \\
    & AudioCLIP$_{\text{ICASSP}22}$ & 20.3 & - & 26.3 & - \\
    \cline{2-6}
    & \cellcolor{azure!10}\textit{\textbf{Ours}:} &  \cellcolor{azure!10} & \cellcolor{azure!10} & \cellcolor{azure!10} & \cellcolor{azure!10} \\
    & \cellcolor{azure!10}$\hookrightarrow$~w/ AV Alignment & \cellcolor{azure!10} \underline{37.2} & \cellcolor{azure!10} \underline{45.6} & \cellcolor{azure!10} \underline{44.1} & \cellcolor{azure!10} \underline{54.6} \\
    & \cellcolor{azure!10}$\hookrightarrow$~w/ AV + LLM Alignment & \cellcolor{azure!10} \textbf{41.1} & \cellcolor{azure!10} \textbf{50.6} & \cellcolor{azure!10} \textbf{48.1} & \cellcolor{azure!10} \textbf{59.5} \\

\bottomrule
    
    \end{tabular}}
\caption{\textbf{Quantitative results on the segmentation test sets.}}
    {
    \label{tab:segmentation_combined}}

\end{table}
If we examine~\Tref{tab:extended} carefully, we observe that our model lags behind prior works only in Ext. Flickr with the LocAcc metric, despite achieving a large performance gap in other metrics. To investigate the reason for this discrepancy, we conduct a further analysis. As in~\Tref{tab:ext_analysis}, our model’s max-F1 score reflects a balanced contribution from both precision and recall, with a gap of 22.37. In contrast, previous methods show much larger gaps -- exceeding 40 -- which indicates that their performance is dominated by recall. This suggests that their max-F1 scores are largely driven by inflated recall, reflecting a tendency to over-detect sounding regions - even in inaudible or mismatched frames. This over-detection behavior is also evident in the precision-recall (PR) curves shown in Figure~\ref{fig:pr_analysis}. While prior methods such as SLAVC~\citep{slavc} and FNAC~\citep{sun2023learning} achieve high recall in the max-F1 metric, they suffer from low precision, resulting in a smaller area under the curve. In contrast, both our \emph{AV Alignment} and \emph{AV + LLM Alignment} variants consistently maintain higher precision at comparable recall levels. This suggests that our model more accurately localizes sounding regions by reducing false activations in silent or irrelevant areas. As shown in Figure~\ref{fig:qualitative_cIoU}, these prior methods often highlight broad visual regions that overlap with the coarsely annotated bounding box and yield high LocAcc scores. However, this does not mean that the model has correctly identified the presence of sound. This discrepancy highlights that high scores on LocAcc neither guarantee accurate detection of sound presence nor reflect true audio-visual understanding.

\subsubsection{Audio-Visual Segmentation} \label{ssec:segmentation}
Our proposed method produces sound source localization outputs that are more compact and fine-grained (see~\Fref{fig:qualitatives}). To further demonstrate the precision of our localization capabilities, we evaluate our method, along with its variants, on the audio-visual segmentation task. This task is particularly suitable for validating fine-grained localization ability, as it requires pixel-level accuracy. Following~\citep{senocak2024aligning}, we use the IS3, VPO-SS, VPO-MS, AVSBench-S4, and AVSBench-M3 datasets, and additionally evaluate on the DenseAV ADE20K dataset. All of these benchmarks provide segmentation masks. These experiments are conducted in a zero-shot setting, where both our model and the compared baselines are trained on VGGSound-144K and directly tested on the target datasets without further fine-tuning. The audio-visual segmentation results are summarized in~\Tref{tab:segmentation_combined}. Consistent with earlier quantitative evaluations, our method achieves superior performance compared to existing approaches; however, the performance gap is even more pronounced in the segmentation setting, with, for instance, a \textcolor{PastelGreen}{31.7\%} cIoU improvement on the S4 dataset. This is expected, as our model tends to generate pixel-level accurate localization maps, in contrast to the coarser, blob-shaped outputs typical of previous methods. Moreover, we observe that the \emph{AV + LLM Alignment} variant consistently provides additional performance gains across all evaluated datasets. Some qualitative results are also presented in~\Fref{fig:qualitatives}.
\begin{table}
\centering
\resizebox{1.0\linewidth}{!}{
\setlength{\tabcolsep}{3pt}
\begin{tabular}{clcccc}
\toprule
& \textbf{Method} & \textbf{IIoU } & \textbf{w/ Adap. } & \textbf{IAUC } & \textbf{w/ Adap. } \\
\midrule
\multirow{16}{*}{\rotatebox[origin=c]{90}{\textbf{IS3}}}
& LVS$_{\text{CVPR}21}$  & 6.5 & 11.2 & 26.0 & 25.3 \\
& EZ-VSL$_{\text{ECCV}22}$  & 7.4 & 13.0 & 26.4 & 26.6 \\
& SSL-TIE$_{\text{ACM MM}22}$  & 9.4 & 19.0 & 28.4 & 31.5 \\
& SLAVC$_{\text{NeurIPS}22}$  & 7.5 & 14.5 & 26.3 & 25.5 \\
& MarginNCE$_{\text{ICASSP}23}$  & 11.5 & 23.7 & 29.4 & 32.5 \\
& FNAC$_{\text{CVPR}23}$  & 11.5 & 22.4 & 28.9 & 31.0 \\
& Alignment$_{\text{ICCV}23}$  & 15.8 & 37.6 & 31.4 & 39.5 \\
\cline{2-6}
& \textit{Baselines:} \\
& \cellcolor{lightgray!25}\includegraphics[height=1em]{Figures/oracle.png}(\textit{\textbf{Oracle}}) CLIPSeg (w/ GT Text) & \cellcolor{lightgray!25}54.7 & \cellcolor{lightgray!25}78.4 & \cellcolor{lightgray!25}47.3 & \cellcolor{lightgray!25}60.4 \\
& CLIPSeg (w/ WAV2CLIP Text) & 7.4 & 30.7 & 11.6 & 34.9 \\
& CLIPSeg (w/ CLAP Text) & 16.5 & 49.1 & 18.8 & 44.7 \\
& WAV2CLIP$_{\text{ICASSP}22}$ & 16.6 & - & 22.6 & -\\
& AudioCLIP$_{\text{ICASSP}22}$ & 4.8 & - & 15.6 & - \\
\cline{2-6}
\rowcolor{azure!10}
\cellcolor{white} & \textbf{\textit{Ours:}} & & & & \\
\rowcolor{azure!10}
\cellcolor{white} &$\hookrightarrow$~w/ AV Alignment & \textbf{43.4} & \textbf{62.3} & \textbf{38.9} & \textbf{50.5} \\
\rowcolor{azure!10}
\cellcolor{white} & $\hookrightarrow$~w/ AV + LLM Alignment & \underline{43.2} & \underline{62.1} & \underline{38.4} & \underline{49.9} \\
\midrule
\multirow{16}{*}{\rotatebox[origin=c]{90}{\textbf{VPO-MS}}}
& LVS$_{\text{CVPR}21}$  & 21.2 & 24.2 & 24.8 & 27.0 \\
& EZ-VSL$_{\text{ECCV}22}$  & 20.9 & 25.4 & 25.4 & 28.3 \\
& SSL-TIE$_{\text{ACM MM}22}$  & 23.8 & 30.6 & 26.4 & 31.0 \\
& SLAVC$_{\text{NeurIPS}22}$  & 22.4 & 28.4 & 25.8 & 29.3 \\
& MarginNCE$_{\text{ICASSP}23}$  & 24.9 & 28.1 & 26.4 & 29.9 \\
& FNAC$_{\text{CVPR}23}$  & 24.9 & 29.7 & 26.8 & 30.1 \\
& Alignment$_{\text{ICCV}23}$  & 24.9 & 30.5 & 26.6 & 30.9 \\
\cline{2-6}
& \textit{Baselines:} \\
& \cellcolor{lightgray!25}\includegraphics[height=1em]{Figures/oracle.png}(\textit{\textbf{Oracle}}) CLIPSeg (w/ GT Text) & \cellcolor{lightgray!25}47.5 & \cellcolor{lightgray!25}65.0 & \cellcolor{lightgray!25}39.8 & \cellcolor{lightgray!25}53.6 \\
& CLIPSeg (w/ WAV2CLIP Text) & 8.4 & 32.8 & 10.9 & 32.9 \\
& CLIPSeg (w/ CLAP Text) & 26.9 & \underline{50.5} & 25.6 & \underline{43.2} \\
& WAV2CLIP$_{\text{ICASSP}22}$  & 16.8 & - & 19.7 & -\\
& AudioCLIP$_{\text{ICASSP}22}$ & 21.6 & - & 24.5 & - \\
\cline{2-6}
\rowcolor{azure!10}
\cellcolor{white} & \textbf{\textit{Ours:}} & & & & \\
\rowcolor{azure!10}
\cellcolor{white} & $\hookrightarrow$~w/ AV Alignment & \underline{28.8} & 42.2 & \underline{28.4} & 38.4 \\
\rowcolor{azure!10}
\cellcolor{white} & $\hookrightarrow$~w/ AV +  LLM Alignment & \textbf{37.4} & \textbf{51.9} & \textbf{33.7} & \textbf{43.9} \\
\bottomrule
\end{tabular}}
\caption{\textbf{Interactive sound source localization results.} All models are trained on VGGSound-144K dataset.}
\label{tab:interactive_table}
\end{table}
\begin{figure}[t]
    \centering
    \includegraphics[width=\linewidth]{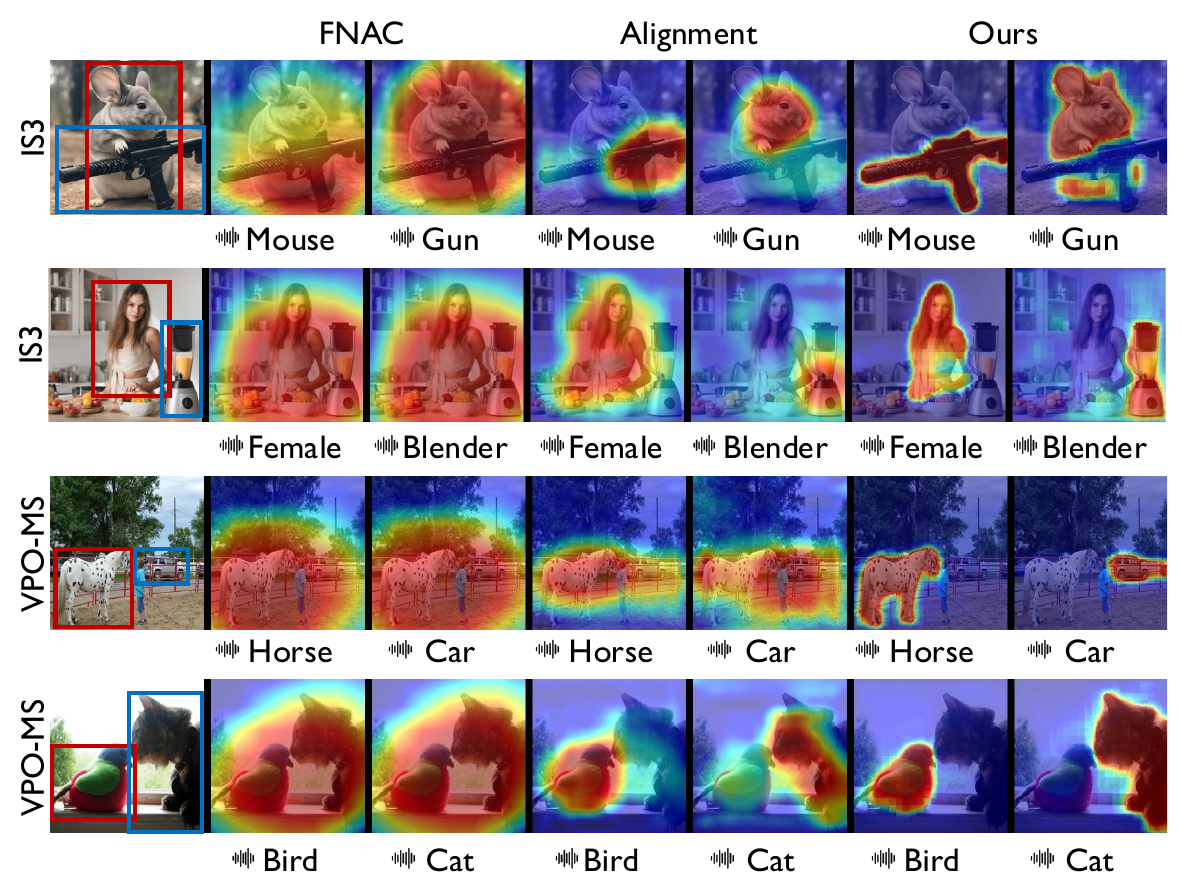}
    \caption{{\bf Qualitative results for interactive sound source localization.
    } 
    Our model interactively perform accurate localization for given different sounds, whereas FNAC results are remain unchanged.}
    \label{fig:interactive_supple}
\end{figure}
\subsubsection{Interactive Sound Source Localization} \label{ssec:interactive}
The desired outcome for sound source localization models is to accurately pinpoint objects that correlate with the sound, rather than simply focusing on prominent or salient objects. Senocak \etal reveal in~\citep{senocak2023alignment,senocak2024aligning} that the majority of existing sound source localization methods fail to localize objects interactively, despite their strong performance in single sound source localization tasks. This task assesses a model’s ability to adjust the localized region within an image when the same image is paired with different sounds present in the scene. For this evaluation, we use the IS3 and VPO-MS benchmarks in a zero-shot setting. The interactive sound source localization capabilities of our model, compared with existing methods, are presented in~\Tref{tab:interactive_table}. Our method outperforms all prior works and baselines by a large margin, consistent with previous experiments. This further demonstrates that our model learns strong audio-visual correspondence by effectively building on CLIP’s cross-modal alignment.

\newpara{Qualitative Results.}~\Fref{fig:interactive_supple} visualizes the results of the Interactive Localization task, comparing our method with recent state-of-the-art approaches, FNAC~\citep{sun2023learning} and Alignment~\citep{senocak2024aligning}, on the IS3 and VPO-MS datasets. Both our method and the Alignment approach demonstrate interactive localization capabilities, but ours yields more accurate results, whereas the FNAC method fails with largely unchanged outputs.

\subsubsection{Sound Source Localization in Mixtures}\label{ssec:multi}
Until now, all tasks presented in this paper have focused on the single sound source localization scenario. However, another important research direction in visual sound source localization is multi-source sound localization, where multiple sound-emitting objects are present in a scene. In this section, we demonstrate the capability of our proposed method to handle this setting too.

\newpara{Datasets and evaluation metrics.} Following AVGN~\citep{mo2023audio} and T-VSL~\citep{mahmud2024t}, we use the VGGSound-Instruments~\citep{hu2022mix} and VGGSound-Duet~\citep{mo2023audio} datasets for evaluation, along with the metrics defined in~\citep{hu2022mix}: CAP (Class-Aware Average Precision), PIAP (Permutation-Invariant Average Precision), cIoU (Class-Aware Intersection over Union), and AUC (Area Under the Curve). Refer to the original papers for detailed metric definitions. We note that prior works report results using cIoU@10 for VGGSound-Instruments and cIoU@30 for VGGSound-Duet. However, these thresholds -- particularly 10\% -- represent relatively small overlapping ratios, which may be insufficient for robust evaluation. Therefore, in addition to the original settings, we also evaluate our models using cIoU@10, cIoU@30 and cIoU@50 on both datasets, alongside other methods with publicly available models.

\newpara{Results.} Unlike existing methods, which are individually trained on each target dataset for multi-source localization with audio mixtures, we directly apply our approach in a zero-shot manner -- just as we have done in previous tasks and experiments. Prior works such as AVGN and T-VSL incorporate additional supervision, including class labels (\eg, as class tokens in visual transformers or as category-level text inputs), during both training and inference. Following the protocol of T-VSL, we use a predefined category list and feed these categories into our grounding module, $G$, which is capable of conditioning on text input.

\begin{figure}[t]
    \centering
    \includegraphics[width=\linewidth]{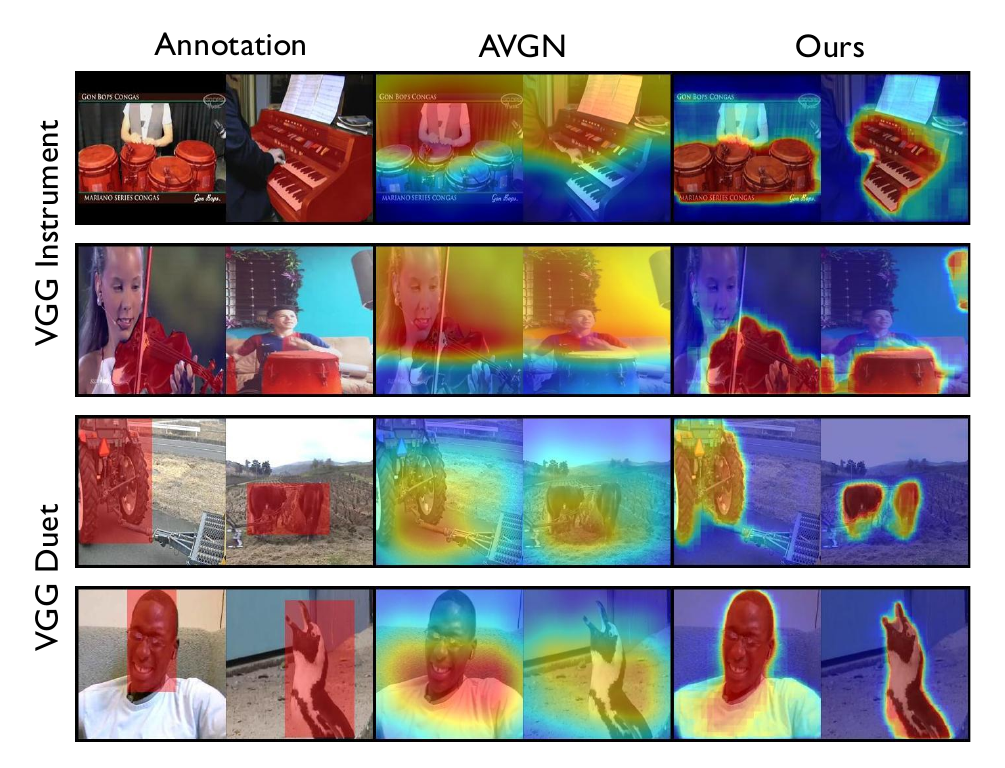}
    \caption{{\bf Qualitative results for multi-source localization.
    } }
    \label{fig:multi}
\end{figure}

Given $N$ classes, our grounder outputs a sounding region mask for each class category, denoted as $\mathbf{M}^G_{n}$, where $n \in {1, \dots, N}$. Following the inference procedure described in~\Sref{ssec:inference}, these masks are subsequently passed to $MaskGen_I$ to obtain the corresponding image-level visual features, denoted as $\boldsymbol{v}^I_n$. Once the visual features are obtained, we estimate the cosine similarity between the audio-driven embedding, $\mathbf{A}$, and each of the $n$ candidate localization regions. Finally, we select the top-$k$ heatmaps corresponding to the highest similarities, and report these as our multi-source localization results. We present the results in~\Tref{tab:multissl}. Notably, although our model is not specifically trained for the multi-source sound localization task, our proposed zero-shot setting outperforms methods that are explicitly trained for it in both datasets. This ability can be attributed to two key factors: (1) the versatility of our approach in seamlessly incorporating text input when available, and (2) the semantically rich audio-driven embeddings, which are highly aligned with CLIP’s visual features. We also provide qualitative results in~\Fref{fig:multi}.
\begin{table}
\centering
\resizebox{1.0\linewidth}{!}{
\setlength{\tabcolsep}{3pt}
\begin{tabular}{clcccccc}
\toprule
& \textbf{Method} & \textbf{CAP} & \textbf{PIAP} & \textbf{AUC} & \textbf{cIoU@10} & \textbf{cIoU@30} & \textbf{cIoU@50} \\
\midrule
\multirow{7}{*}{\rotatebox[origin=c]{90}{\textbf{VGG Inst.}}}
& OTS$_{\text{ECCV}18}$ & 23.3 & 37.8 & 11.7 & 51.2 & - & - \\
& Mix-and-Localize$_{\text{CVPR}22}$ & 21.5 & 37.5 & 15.6 & 73.2 & - & - \\
& AVGN$_{\text{CVPR}23}$ & 27.3 & 42.8 & 18.2 & 77.5 & 21.4 & 5.71 \\
& NoPrior$_{\text{CVPR}24}$ & - & - & - & - & - & -  \\
& T-VSL$_{\text{CVPR}24}$ & 41.8 & - & 31.5 & 89.6 & - & - \\
\cline{2-8}
& \cellcolor{azure!10} \textbf{\textit{Ours:}} & \cellcolor{azure!10} & \cellcolor{azure!10} & \cellcolor{azure!10} & \cellcolor{azure!10} & \cellcolor{azure!10} & \cellcolor{azure!10} \\
&\cellcolor{azure!10} $\hookrightarrow$~w/ AV Alignment & \cellcolor{azure!10} \textbf{64.1} & \cellcolor{azure!10} \textbf{78.9} & \cellcolor{azure!10} \underline{51.6} & \cellcolor{azure!10} \textbf{97.2} & \cellcolor{azure!10} \underline{81.6} & \cellcolor{azure!10} \underline{55.6} \\
&\cellcolor{azure!10} $\hookrightarrow$~w/ AV + LLM Alignment & \cellcolor{azure!10} \underline{63.4} & \cellcolor{azure!10} \underline{78.4} & \cellcolor{azure!10} \textbf{53.1} & \cellcolor{azure!10} \underline{97.0} & \cellcolor{azure!10} \textbf{83.7} & \cellcolor{azure!10} \textbf{57.2} \\
\midrule
\multirow{7}{*}{\rotatebox[origin=c]{90}{\textbf{VGG Duet}}}
& OTS$_{\text{ECCV}18}$ & 10.5 & 12.7 & 15.8 & - & 12.2 & - \\
& Mix-and-Localize$_{\text{CVPR}22}$ & 16.3 & 22.6 & 20.5 & - & 21.1 & - \\
& AVGN$_{\text{CVPR}23}$ & 21.9 & 28.1 & 23.8 & \underline{84.5} & 26.2 & 22.2 \\
& NoPrior$_{\text{CVPR}24}$ & 32.5 & 44.4 & 29.2 & - & 46.9 &  \\
& T-VSL$_{\text{CVPR}24}$ & 35.7 & - & 37.9 & - & 40.1 &  \\
\cline{2-8}
 & \cellcolor{azure!10} \textbf{\textit{Ours:}} & \cellcolor{azure!10} & \cellcolor{azure!10} & \cellcolor{azure!10} & \cellcolor{azure!10} &  \cellcolor{azure!10} & \cellcolor{azure!10} \\
 & \cellcolor{azure!10} $\hookrightarrow$~w/ AV Alignment & \cellcolor{azure!10} \textbf{56.8} & \cellcolor{azure!10} \textbf{63.2} & \cellcolor{azure!10} \textbf{42.2} & \cellcolor{azure!10} \textbf{84.9} & \cellcolor{azure!10} \textbf{62.7} & \cellcolor{azure!10} \textbf{42.6} \\
  & \cellcolor{azure!10} $\hookrightarrow$~w/ AV + LLM Alignment & \cellcolor{azure!10} \underline{53.7} & \cellcolor{azure!10} \underline{61.5} & \cellcolor{azure!10} \underline{38.7} & \cellcolor{azure!10} 77.7 & \cellcolor{azure!10} \underline{55.9} & \cellcolor{azure!10} \underline{37.2} \\
\bottomrule
\end{tabular}}
\caption{\textbf{Performance comparison of multi-source localization.}}
\label{tab:multissl}
\end{table}

\subsection{Ablation Results}\label{ssec:ablation}
\subsubsection{Ablation on different combinations of loss functions} Our method is optimized by a combination of three loss functions, \ie $ACL_I$, $ACL_F$, and area regularization. Here, we perform ablation experiments to understand the impact of each loss function. We primarily conduct experiments on VGG-SS, AVS-Bench S4 and Extended VGG-SS datasets. Results are in~\Tref{tab:ablation}.
\begin{table}[ht]
    \centering
    \resizebox{1.0\linewidth}{!}{
    \setlength{\tabcolsep}{3pt}
    \begin{tabular}{lccccccccc}
    \toprule
    &\multicolumn{3}{c}{} & \multicolumn{2}{c}{\textbf{VGG-SS}} & \multicolumn{2}{c}{\textbf{AVS (S4)}} & \multicolumn{2}{c}{\textbf{Ext. VGG-SS}} \\
    \textbf{} & $ACL_I$ & $ACL_F$ & $Reg$ & \textbf{cIoU} & \textbf{AUC} & \textbf{mIoU} & \textbf{F-score} & \textbf{AP} & \textbf{max-F1} \\ \midrule
    (A)  & \ding{51} & \ding{55} & \ding{55} & 40.42 & 40.84 & 38.55 & 45.94 & 28.59 & 35.90  \\
    (B)  &  \ding{55} & \ding{51} & \ding{55} & 2.30 & 7.46 & 4.08 & 22.59 & 0.86 & 1.80  \\
    (C)  & \ding{51} & \ding{51} & \ding{55} & 46.61 & 44.71 & 53.06 & 63.01 & 40.72 & 47.90  \\
    (D)  & \ding{51} & \ding{55} & \ding{51} & 41.08 & 41.01 & 41.93 & 48.99 & 33.37 & 41.30  \\
    (E)  & \ding{55} & \ding{51} & \ding{51} & 35.15 & 38.36 & 32.06 & 41.05 & 39.91 & 47.20  \\
    \rowcolor{azure!10}
    (F)  & \ding{51} & \ding{51} & \ding{51} & \textbf{49.46} & \textbf{46.32} & \textbf{59.76} & \textbf{69.03} & \textbf{40.79} & \textbf{49.10}  \\    
    \toprule
    \end{tabular}}
    \caption{\textbf{Ablative experiments on our method by using different combinations of loss functions.}}
    \label{tab:ablation}
\end{table}
\begin{figure}[tp]
    \centering
    \includegraphics[width=\linewidth]{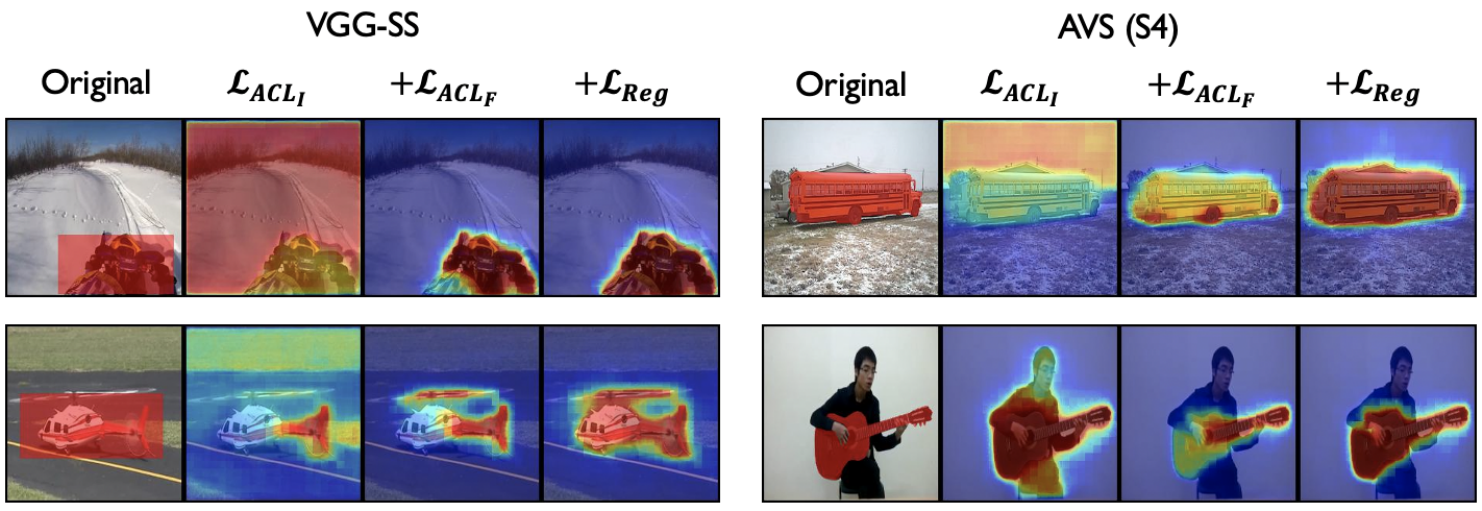}
    \caption{{\bf Sound localization results by using different combinations of loss functions.} }
    \label{fig:qualitative_ablation}
\end{figure}

As revealed by results (A) and (B), using $ACL_I$ is crucial to enable our model to learn the corresponding audio-visual alignment. On the other hand, relying solely on $ACL_F$ is not effective for learning audio-visual alignment, as it primarily focuses on suppressing unrelated areas. However, as demonstrated by the results of (A \emph{vs.} C) and (B \emph{vs.} C), the combination of these two loss functions is complementary. As mentioned earlier, $ACL_F$ contributes to performance enhancement by suppressing background areas. Furthermore, the results from the experiments (C \emph{vs.} F) show that area regularization provides additional improvements by constraining the size of the activated regions. 

\newpara{Visualization of the ablation experiments.} The visual results are presented in~\Fref{fig:qualitative_ablation}. As demonstrated, when using only $ACL_I$, we observe that background areas remain activated (also discussed in~\Sref{ssec:alignment}). As evident in the third column, the addition of $ACL_F$ helps eliminate the background pixels (non-sounding areas). However, it is noticeable that the outputs of $ACL_I$+$ACL_F$ can be relatively less completed. With the area regularizer, the final output of our model becomes more complete and fine-grained.

\begin{table}[t]
    \centering
    \resizebox{1.0\linewidth}{!}{
    \setlength{\tabcolsep}{3pt}
    \begin{tabular}{lcccccc}
    \toprule
     Caption & \multicolumn{2}{c}{\textbf{VGG-SS}} & \multicolumn{2}{c}{\textbf{AVS (S4)}} & \multicolumn{2}{c}{\textbf{Ext. VGG-SS}} \\
     Modality & \textbf{cIoU} & \textbf{AUC} & \textbf{mIoU} & \textbf{F-Score} & \textbf{AP} & \textbf{max-F1} \\ \midrule
     AV Align. & \underline{49.5} & \underline{46.3} & 59.8 & \underline{69.0} & 40.8 & 49.1 \\
     \midrule
    Audio & 46.7 & 43.3 & 58.3 & 66.2 & 42.8 & 47.4 \\
    Vision & 49.0 & 45.0 & \underline{60.7} & 68.3 & \underline{44.3} & \underline{49.4} \\
    \rowcolor{azure!10}
    Audio + Vision & \textbf{52.1} & \textbf{46.9} & \textbf{61.8} & \textbf{69.6} & \textbf{46.7} & \textbf{51.9} \\
    \toprule
    \end{tabular}}
    
    \caption{\textbf{Ablation on caption modalities for caption alignment variant.}}
    \label{tab:ablation_caption}
\end{table}

\subsubsection{Ablation on different modality captions for LLM-based guidance alignment} Our \emph{AV + LLM Alignment} variant uses the output of an LLM as auxiliary self-supervision, where the LLM’s response is generated based on obtained captions from vision and audio samples. To see the impact of each caption modality, we trained our model by providing the LLM with captions from only a single modality. We present the results in~\Tref{tab:ablation_caption}.

We observe that captions derived from the vision modality yield better performance than those from the audio modality. However, incorporating single-modality captions along with their corresponding LLM responses does not lead to improved performance; in fact, they underperform compared to the AV Alignment variant. This can be attributed to the fact that single-modality captions prompt the LLM to infer object-aware audio-visual understanding using only one modality, which may be insufficient to accurately describe the scene or identify the sound-emitting object. As a result, this weakens the audio-visual correspondence the model is learning between audio-driven embeddings and visual features. In contrast, when captions from both modalities are provided to the LLM, the response is likely to be more accurate and semantically aligned with the scene. This enriched multi-modal supervision enables the model to surpass the performance of the AV Alignment variant alone.

\begin{figure*}[tp]
    \centering
    \includegraphics[width=\linewidth]{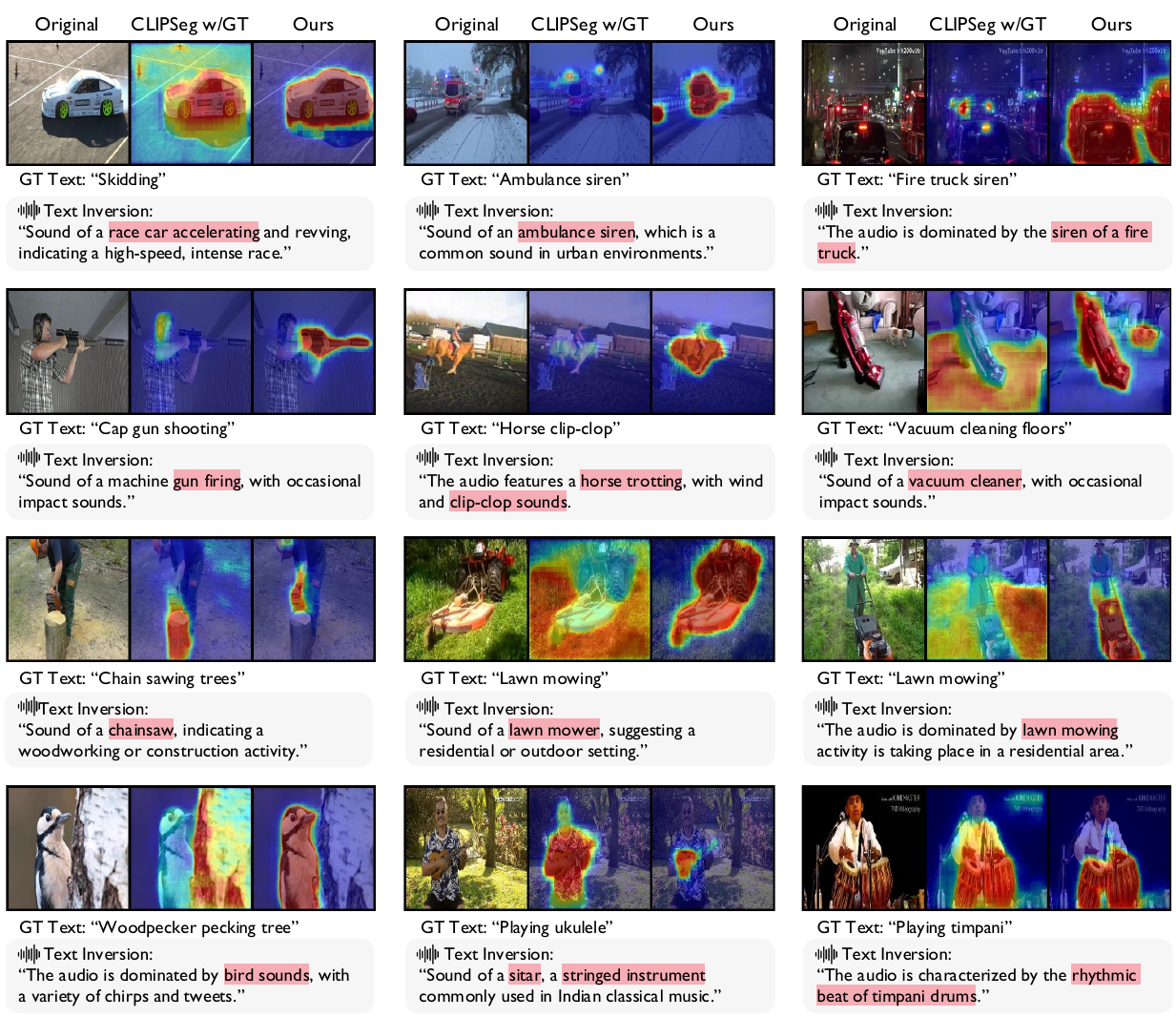}
    \caption{{\bf Analysis on AudioTokenizer and visual comparison with \includegraphics[height=1em]{Figures/oracle.png} Oracle method.} Sound sources cannot always be properly localized, even when class-label text information is used ((\textit{\textbf{Oracle}}) CLIPSeg (w/ GT Text)). However, our model can localize sound sources accurately using only the audio embeddings learned through audio-visual correspondence. This indicates that our AudioTokenizer enables the generation of semantically accurate audio embeddings. We also visualize what our audio embeddings describe using text inversion approach.}
    \label{fig:oracle_comparison}
\end{figure*}

\subsection{Further Analysis on Audio Tokenizer}\label{ssec:tokenizeranalysis}
In this section, we analyze the capabilities of our AudioTokenizer and audio-driven embeddings, focusing on their ability to describe audio samples in a richer and more detailed manner. We also provide qualitative examples to emphasize that the \includegraphics[height=1em]{Figures/oracle.png} Oracle method (CLIPSeg w/ GT text) method is not universally applicable and has inherent limitations. Throughout the paper, we have discussed that the sound source localization task is inherently unlabeled, and that ground-truth class information for benchmark samples is not always available, making the Oracle method a hypothetical reference. To further demonstrate why solving sound source localization through text input-based segmentation approaches is problematic, we present qualitative examples showing that class labels are often insufficiently descriptive to accurately localize sound sources. In contrast, learning audio-visual alignment equips the audio encoder with greater descriptive power, enabling our model -- enhanced by the proposed AudioTokenizer -- to represent audio more richly and localize sound sources more effectively. Note that this analysis is based on our \emph{AV Alignment} variant, and all examples are drawn from the VGG-SS dataset.

We present qualitative comparison results on the VGG-SS dataset in~\Fref{fig:oracle_comparison}. These results highlight two key observations: (1) Sound sources cannot always be accurately localized, even when class label text information is available. Audio provides a broader descriptive context in visual scenes, and sound source localization models should focus on learning true audio-visual correspondence. (2) The successful results of our model compared to the Oracle method demonstrates that our AudioTokenizer module effectively encodes the true audio context, enabling proper learning of the audio-visual correspondence objective.

To further support our claim that the audio-driven embeddings effectively capture audio content, we apply a text inversion approach with them to visualize what they represent and how they describe the input. We perform the text inversion experiment using a CLIP-based captioning model. Specifically, we use CAPDEC~\citep{nukrai2022text}, an autoregressive model that generates text by conditioning a GPT2~\citep{radford2019language} decoder on CLIP embeddings. We train CAPDEC on image-text pairs constructed from the VGGSound dataset by generating audio captions via GAMA~\citep{ghosh2024gama} and pairing them with the corresponding images. Once trained, the model receives our audio-driven embeddings instead of image embeddings, allowing us to generate descriptive text from audio to see the semantics encoded by the learned audio representations. Text inversion results are also in~\Fref{fig:oracle_comparison}. The results demonstrate that our audio-driven embeddings carry accurate and descriptive semantic information.

\section{Conclusion}
In this work, we introduced a self-supervised framework for sound source localization that leverages the multimodal alignment knowledge of foundational models, specifically CLIP. Our AudioTokenizer module transforms audio signals into CLIP-compatible tokens, enabling audio-visual alignment without relying on textual class labels. Through contrastive learning, our model effectively grounds sounding regions and aligns audio and visual representations using audio-visual correspondence. Extensive experiments across diverse tasks -- including single-source and multi-source localization, segmentation, interactive localization, and robustness -- demonstrate that our method consistently outperforms existing approaches, achieving state-of-the-art performance and strong generalization in zero-shot settings. Additionally, we propose an LLM-guided extension to further enhance alignment through object-aware audio-visual scene understanding during training. Our findings suggest that the core challenge of sound source localization -- achieving strong audio-visual alignment -- can be addressed by building upon structured multimodal alignment priors offered by large-scale pre-trained foundation models.
\section{Acknowledgment}
This work was supported by Institute for Information \& communications Technology Planning \& Evaluation (IITP) grant funded by the Korea government (MSIT, RS-2025-02215122) and Electronics and Telecommunications Research Institute (ETRI) grant funded by the Korean government (25ZC1100, The research of the basic media$\cdot$contents technologies).

\noindent\textbf{Data availability statements.}
All data supporting the findings of this study are available online.
The VGGSound dataset can be downloaded from \url{https://www.robots.ox.ac.uk/~vgg/data/vggsound/}. 
The SoundNet-Flickr dataset can be downloaded from \url{https://github.com/ardasnck/learning_to_localize_sound_source}.
The IS3 dataset can be downloaded from \url{https://github.com/kaistmm/SSLalignment}.
The VPO datasets can be downloaded from \url{https://github.com/cyh-0/CAVP}.
The AVSBench datasets can be downloaded from \url{http://www.avlbench.opennlplab.cn/download}.
The DenseAV ADE20K datasets can be downloaded from \url{https://github.com/mhamilton723/DenseAV}.
The Extended VGG-SS/Flickr datasets can be downloaded from \url{https://github.com/stoneMo/SLAVC}.
The VGGSound-Instrument datasets can be downloaded from \url{https://web.eecs.umich.edu/~ahowens/mix-localize/}.
The VGGSound-Duet datasets can be downloaded from \url{https://github.com/stoneMo/AVGN}.

\bibliographystyle{spbasic}      
\bibliography{main}   
\end{document}